\documentclass{article}

\PassOptionsToPackage{numbers,sort&compress}{natbib}

\usepackage[main, final]{neurips_2026}

\usepackage[utf8]{inputenc} 
\usepackage[T1]{fontenc}    
\usepackage{hyperref}       
\usepackage{url}            
\usepackage{booktabs}       
\usepackage{amsfonts}       
\usepackage{nicefrac}       
\usepackage{microtype}      
\usepackage{xcolor}         

\usepackage{setspace}
\usepackage{multirow} 
\usepackage{graphicx}
\usepackage{subcaption}
\usepackage{amsmath} 
\usepackage{cleveref}
\usepackage{xspace}
\usepackage{tikz}
\usepackage{amssymb}
\usetikzlibrary{arrows.meta,calc,fit}
\newcommand{\BA}{BA\xspace}
\newcommand{\BAR}{BA$_{\mathrm{R}}$\xspace}
\newcommand{\PSNRW}{PSNR$_{\mathrm{W}}$\xspace}
\newcommand{\PSNRR}{PSNR$_{\mathrm{R}}$\xspace}
\newcommand{\dPSNR}{$\Delta$PSNR\xspace}

\usepackage{placeins}

\crefname{algocf}{Algorithm}{Algorithms}
\usepackage[ruled,vlined]{algorithm2e}
\title{FiLM-CAM: Keyed Feature Modulation for Conditional-Access Watermarking}

\author{%
  Yash Kulthe$^{1,2}$\quad
  Shruti Agarwal$^{2}$\quad
  Vishal Asnani$^{2}$\\
  \textbf{Andrew Gilbert}$^{1}$\quad
  \textbf{John Collomosse}$^{1,2}$\\[4pt]
  $^{1}$DECaDE, University of Surrey, UK\qquad
  $^{2}$Adobe Research\\[2pt]
  \texttt{\{y.kulthe, a.gilbert, j.collomosse\}@surrey.ac.uk}\\
  \texttt{\{shragarw, vasnani, collomos\}@adobe.com}
}

\begin{document}

\maketitle

\begin{abstract}
  We propose FiLM-CAM; a conditional-access image watermarking approach that enforces authorized encoding, decoding, and removal through keyed feature modulation. A secret key conditions a Feature\allowbreak-wise Linear Modulation (FiLM) mechanism that transforms the internal representation of the watermark signal before embedding, thereby coupling the payload to the key at the feature, rather than pixel level. The same key must be presented to condition the decoder to correctly recover the payload embedded in an image.  To enforce conditional access, we train the system to maximize the performance gap between authorized and unauthorized use, ensuring reliable extraction under the correct key while inducing failure under incorrect keys. In addition, we design our encoder to estimate the watermark residual invariant to the presence of pre-existing watermarks. This design enables conditional watermark removal via simple re-encoding and subtraction of the residual using iterative refinement, contingent on knowledge of the secret key.  We evaluate FiLM-CAM across multiple watermarking encoder–decoder backbones, demonstrating that keyed feature modulation serves as an effective, architecture-agnostic conditional access module (CAM) for image watermarking.
\end{abstract}

\vspace{-0.5em}
\section{Introduction}
\label{sec:intro}
\vspace{-0.5em}
Generative AI raises societal concerns over media integrity, given the speed and scale at which synthetic or misleading imagery can be produced and disseminated. Invisible watermarking has emerged as a practical mitigation, embedding imperceptible signals into image content that either flag the presence of AI-generated content (AIGC) \cite{gowal2025synthidimage,fernandez2023stable} or enable content re-identification to establish provenance for media authenticity \cite{bui2025trustmark,collomosse2024beyond}, rights \cite{Balan-CVMP-2023,Balan-WMF-2023} and licensing \cite{Collomosse-IEEECGA-2026}.

It is not always desirable for such watermarking to be openly accessible. Organizations may wish to control (or `condition') access to the encoding and/or decoding of watermarks to protect against spoofing, unauthorized removal, or monitoring.  This has led to a proliferation of private watermarking encoder–decoder models, each trained independently for an individual organization, imposing a deployment burden on implementers who must distribute and maintain multiple private decoders. It would be more efficient to train a watermarking framework in which access control is intrinsic to the model’s operation i.e., where encoding, decoding, and removal are functionally contingent on possession of an authorized key, without requiring separate models for each private deployment. 

This paper proposes {\bf FiLM-CAM}\footnote{Project page: \url{https://yash-research.github.io/filmcam/}}, a conditional-access watermarking method that enforces authorized encoding, decoding, and removal of image watermarks through keyed feature modulation. Rather than binding the secret solely to the watermark signal, FiLM-CAM integrates the key into the encoder–decoder network's internal feature hierarchy via Feature-wise Linear Modulation (FiLM). The key modulates intermediate activations, gating how watermark information is represented, embedded, and subsequently extracted. As a result, correct decoding requires not only the image but also condition ing by the valid key, while unauthorized keys induce systematic failure. Our technical contributions are:

\noindent {{\bf 1. Conditional Watermarking via Keyed Feature Modulation}. We introduce a Conditional Access Module (CAM) that embeds access control directly into a watermarking encoder–decoder through FiLM-based feature modulation \cite{perez2018film}. A secret key conditions intermediate representations, enabling a single shared backbone to support multiple private watermarking domains while enforcing a performance gap between authorized and unauthorized keys. } 

\noindent {{\bf 2. Logit-Space Embedding with Conditional Residual Removal.} 
We formulate watermark embedding in the logit space rather than via direct addition of pixel residuals. The encoder is trained to predict a key-conditioned residual invariant to the presence or absence of the watermark, enabling removal by subtracting the residual. A gain-controlled refinement step yields high-fidelity removal (PSNR $\sim$ 54dB) when the correct key and message are provided.}
\vspace{-0.5em}
\vspace{-0.5em}

\vspace{-0.5mm}
\section{Related Work}
\vspace{-0.5em}
Image watermarking has been widely applied to tasks including leak detection \cite{cox2007digital}, forensic detection of manipulated or AI-generated media \cite{asnani2022proactive, gowal2025synthidimage}, data attribution \cite{asnani2024promark,asnani2025custommark}, and content provenance \cite{videosealarxiv,bui2025trustmark}.

\noindent \textbf{Classical image watermarking} methods embedded signals directly into pixels via Least Significant Bit (LSB) encoding \cite{vanschyndel1994digital}. Subsequent frequency-domain approaches include DCT and DWT-based embedding \cite{barni1998dct,barni2001dwt}. Hybrid methods such as DWT–DCT–SVD \cite{navas2008dwt} further improved robustness under compression and remain widely used for AIGC labelling. Spread-spectrum watermarking \cite{cox1997secure} later improved robustness by using pseudo-random embedding patterns and offered conditional access via shared seeds that control frequency patterns. More recently, data-hiding approaches have demonstrated reversible watermarking, enabling recovery of the original image after watermark removal \cite{chen2025learning}.

\begin{figure}[t]
  \centering
  \input{figure1_tikz}
  \caption{Architecture of FiLM-CAM. \textbf{Stage 1}: The Embedder generates watermark residuals, which are added to the cover image, to get a watermarked image. The Extractor recovers the watermark from the image if it receives the correct key and an all-zero bit message in case of an incorrect key. \textbf{Stage 2}: The Embedder network is trained to output residuals when given a watermarked image, such that it can be used to remove the watermark. \textbf{Conditional Access Module (CAM)}: A binary key $k$ is mapped to an embedding $e_k=\phi(k)$ and transformed via $\psi$ to produce feature-wise parameters $(\gamma_k,\beta_k)$. These condition the intermediate feature maps via FiLM modulation $F_{\text{cond}}=\gamma_k \odot F + \beta_k$, enabling key-dependent watermark embedding, decoding, and removal.}
  \label{fig:arch}
\end{figure}

\noindent \textbf{Deep Learning methods} jointly learn encoder–decoder (E-D) architectures to embed payloads into images.  Early E-D models such as HiDDEN \cite{zhu2018hidden}  evolved into StegaStamp \cite{tancik2020stegastamp} and StampOne \cite{shadmand2024stampone}, which improve the geometric resilience for print; SSL \cite{fernandez2022watermarking} and RoSteALS \cite{bui2023rosteals}, which operate in latent space; and TrustMark \cite{bui2025trustmark}, which introduced point convolutions and a residual scaling trick to mitigate the low-resolution of E-D models.  SSL\cite{fernandez2022watermarking} requires shared knowledge of a vector parameter between the encoder and decoder. More recent work explores adversarial training of watermarking networks, such as PixelSeal \cite{souvcek2025pixel}, and transformer-based decoders in WAM \cite{sander2024watermark} and VideoSeal \cite{videosealarxiv}, the latter of which demonstrates temporal coherence across frames for video watermarking. Whilst these approaches operate post-hoc on existing imagery, complementary work embeds watermarks implicitly during the synthesis of generative AI images \cite{fernandez2023stable,wen2023treerings}; in Tree-Ring \cite{wen2023treerings}, the key is a pattern injected into the initial diffusion noise, effectively the watermark itself. Key-based FNNS explores conditional access for generation time watermarking in \cite{luo2023securing}. Conditional access post-hoc watermarking for audio was explored very recently \cite{xu2025wake}. LatentSeal \cite{evennou2025latentseal} secures a semantic (text-derived) watermark payload with a secret, key-seeded rotation applied at inference, while the watermarking network itself remains key-agnostic. Adaptation of image watermarking models using LoRA was recently explored in \cite{Saberi-WACV-2027}. In contrast to methods where the key defines the watermark pattern (Tree-Ring) or transforms the payload (LatentSeal), FiLM-CAM uses the key to condition the network itself. To our knowledge, no prior learned E-D method has explored conditional-access post-hoc image watermarking via a single model. FiLM-CAM addresses this gap by conditioning watermark embedding and decoding through feature-wise modulation, enabling a shared encoder–decoder backbone whose behaviour varies with the supplied key. 

\noindent \textbf{Watermark removal methods} aim to remove embedded signals while preserving image quality, often in a blind manner. Classical removal attacks rely on signal processing operations such as filtering, compression, or geometric transformations to degrade watermark decoding \cite{cox2007digital}, but they often also degrade image quality. Learned removal approaches have recently been explored in the context of robust watermarking systems \cite{bui2025trustmark}, via separate removal networks trained to estimate and subtract watermark perturbations. Other approaches treat removal as an adversarial objective, using optimization or generative models to destroy the watermark and reconstruct a plausible image \cite{zhao2024invisible,liu2024ctrlregen}. However, such approaches may produce reconstructions that deviate from the original image or fail against strong watermarking schemes. In contrast, FiLM-CAM incorporates watermark removal directly within a single E–D watermarking architecture: a single embedder network is trained to estimate the watermark residual under the correct key, enabling controlled watermark removal through subtraction while preserving high-quality reconstruction of the original image.

\vspace{-2mm}
\section{Methodology}
\label{sec:methodology}
\vspace{-2mm}

FiLM-CAM is a conditional-access neural watermarking method that gates embedding, decoding, and removal on possession of a secret key (\Cref{fig:arch}). Given a cover image $x$, payload $m$, and key $k$, an embedder network predicts a watermark residual that is composed with the image to produce a watermarked output $x_{w}$. An extractor network then attempts to recover the embedded payload from $x_{w}$, but only when conditioned on the correct key. When incorrect keys are used, decoding is encouraged to collapse to an invalid output. For watermark removal without introducing a separate remover network, FiLM-CAM reuses the embedder. The embedder predicts the corresponding watermark perturbation, which can be subtracted in the same transformed domain used for embedding.

\vspace{-2mm}
\subsection{Conditional Access Module (CAM)}
\label{sec:CAM}
\vspace{-2mm}
The method's conditional behavior is enabled by a lightweight Conditional Access Module (CAM) implemented using Feature-wise Linear Modulation (FiLM) \cite{perez2018film}, which supports key-dependent watermark embedding, decoding, and removal. CAM injects the key into the intermediate feature representations of both the embedder and the extractor, thereby making the watermarking process representation-level and dependent on the supplied key. Let $k \in \{0,1\}^L$ denote a binary key of length $L$. CAM first maps $k$ into a continuous embedding vector $e_k = \phi(k)$, where $\phi(\cdot)$ is a learned key embedding network. In our implementation, $\phi$ is a lightweight three-layer MLP that projects the binary key into a $d$-dimensional latent space. This embedding is shared across the embedder and extractor, so that both operate in a common key-conditioned feature space.

Given the key embedding $e_k$, CAM predicts a pair of channel-wise modulation parameters, $\gamma_k$ and $\beta_k$, using the projection $[\gamma_k, \beta_k] = \psi(e_k)$, where $\psi(\cdot)$ is a linear layer which maps the key embedding to feature-wise scale and shift terms. For an intermediate feature tensor $F$, FiLM conditioning is then applied as: $\mathrm{CAM}(F, k) = \gamma_k \odot F + \beta_k,$ where $\odot$ denotes the Hadamard product. For convolutional feature maps, $\gamma_k$ and $\beta_k$ are broadcast across the spatial dimensions.

In the embedder, CAM is applied to the bottleneck representation of the residual-predicting encoder, allowing the key to influence the latent features from which the watermark residual is generated. In the extractor, CAM is applied to the high-level image representation immediately before message prediction, making correct decoding contingent on the supplied key. As a result, the correct key aligns the embedding and extraction processes, while an incorrect key disrupts this alignment and induces decoding failure (c.f. \Cref{sec:sched}). 

\vspace{-0.5em}
\subsection{Watermark Embedder/Extractor}
\vspace{-0.5em}
The watermark embedder $E$ is implemented as a redesigned U-Net-based architecture inspired by PixelSeal\cite{souvcek2025pixel}, with 13.3M parameters. 
Given a resized cover image $x \in [-1,1]^{3 \times 256 \times 256}$, watermark message (secret) $m \in \{0,1\}^n$, and conditioning key $k \in \{0,1\}^L$, the secret $m$ is first projected to a spatial representation and concatenated with the RGB cover image at the input. The resulting tensor is processed by an encoder-decoder network (the Embedder $E$) with skip connections and a residual bottleneck, producing a residual $r$: $r = E(x, m, k); \ \  r \in \mathbb{R}^{3 \times 256 \times 256},$

This residual is then combined with the cover image through the composition function of \Cref{sec:watermark_composition} to obtain the final watermarked image $x_{w}$: $x_{w} = \mathcal{C}(x, r),  \ \ x_{w} \in [-1,1]^{3 \times 256 \times 256},$
where $\mathcal{C}(\cdot)$ denotes the watermark composition operator.  FiLM
modulation is applied at the bottleneck representation so that the residual depends jointly on the image, message, and key. 

The watermark extractor $X$ receives a watermarked image and predicts the embedded message $\hat{m}$ conditioned on the supplied key: $\hat{m} = X(x_{w}, k), \ \ \hat{m} \in \{0,1\}^n$.
We use ConvNeXt-Base \cite{liu2022convnet} as the extractor backbone, replacing its prediction layer with an $n$-dimensional output head for bit prediction. The input image is passed through the visual backbone and globally pooled to obtain a compact feature representation. FiLM conditioning is then applied immediately before the final prediction layer. In our experiments, both the message length $n$ and key length $L$ are 100 bits, and the key is mapped to a 64-dimensional embedding shared by the embedder and extractor.
\vspace{-0.5em}
\subsection{Watermark Composition}
\label{sec:watermark_composition}
\vspace{-0.5em}

A common way to construct a watermarked image is to directly add the predicted residual to the cover image, i.e., $x_{w} = x + r$, clipping pixels outside the valid range. In our work, we compose the watermark in a transformed logit space rather than directly in the image space. We adopt this formulation primarily for removal: composition in the unconstrained logit domain yields cleaner watermark removal. Empirically, it matches additive composition in embedding quality but improves removal fidelity (\Cref{sec:ablation}).  Given a cover image $x \in [-1,1]^{3 \times 256 \times 256}$ and a predicted residual $r \in \mathbb{R}^{3 \times 256 \times 256}$, we first apply a safety margin $\kappa \in (0,1)$ and map the image to an unconstrained transformed domain $z_0$: $x_{\mathrm{safe}} = \kappa x, \ \ z_0 = \operatorname{atanh}(x_{\mathrm{safe}}).$

The watermark is then added in this transformed space. Composing in this space introduces a saturation issue, which motivates our gain term: because $\tanh(\cdot)$ saturates near $\pm 1$, a na\"ive additive perturbation in $z$-space biases the watermark toward mid-intensities. We compensate with an image-dependent gain term: $g(x) = 1/(1 - x_{\mathrm{safe}}^2 + \epsilon),$
where $\epsilon > 0$ is a small constant for numerical stability. For pixels near zero, $g(x)$ remains close to $1$, so the residual scale is largely preserved. Near the image boundaries, however, $g(x)$ becomes larger, amplifying residuals in regions where the $\tanh(\cdot)$ mapping is less sensitive. The transformed-space perturbation is therefore defined as $\Delta z = \alpha \,\tanh(r) \odot g(x)$ 
where $\alpha > 0$, a global scalar controls the global watermark strength and $g(x)$ is applied element‑wise. The final watermarked image is obtained by adding $\Delta z$ in the transformed space and mapping back to the image domain as $x_{w} = \tanh(z_0 + \Delta z)$.  Combining these expressions gives the complete watermark composition operator: $\mathcal{C}(x, r) = \tanh\!\left(\operatorname{atanh}(\kappa x) + \alpha\,\tanh(r) \odot (1 - (\kappa x)^2 + \epsilon)^{-1}\right).$
\vspace{-0.5em}
\subsection{Noise Layer}
\vspace{-0.5em}
To improve robustness, we insert a noise layer $N$ after watermark composition during training as $\tilde{x}_w = N(x_{w})$.  The noise layer applies one affine and two randomly sampled further transforms to simulate image distortions. Fixed transforms include random horizontal flipping and cropping/resizing, while the two additional transforms are selected from a pool that includes JPEG compression, photometric adjustments, blur, additive noise, posterization, RGB shift, and geometric distortions such as perspective change, resizing, and random cropping.

\begin{algorithm}[t]
\footnotesize   
\setlength{\algomargin}{6pt}   
\SetAlgoNlRelativeSize{-1}     
\SetAlCapSkip{2pt}             

\caption{Invert watermark using iterative refinement}
\label{alg:invert_watermark}

\KwIn{Watermarked image $\mathbf{x_{w}}$, Removal residual $\mathbf{r_{rem}}$, $\kappa=0.95$, Watermark strength $\alpha$, $\epsilon=10^{-3}$, $n_{refine}=20$}
\KwOut{Estimated cover image $\mathbf{c}_{est}$}

$\mathbf{z}_{stego} \gets \text{atanh}(\text{clamp}(\mathbf{x_{w}}, -0.999, 0.999))$\;
$\mathbf{c}_{est} \gets \mathbf{x_{w}}$\;

\For{$i \gets 1$ \KwTo $n_{refine}$}{
    $\mathbf{c}_{safe} \gets \kappa \cdot \mathbf{c}_{est}$\;
    $\mathbf{c}_{safe} \gets \text{clamp}(\mathbf{c}_{safe}, -0.999, 0.999)$\;
    $\mathbf{g} \gets 1.0 / (1.0 - \mathbf{c}_{safe}^2 + \epsilon)$\;
    $\Delta\mathbf{z} \gets \alpha \cdot \text{tanh}(\mathbf{r_{rem}})\cdot \mathbf{g}$\;
    $\mathbf{z}_{cover} \gets \mathbf{z}_{stego} - \Delta\mathbf{z}$\;
    $\mathbf{c}_{est} \gets \text{tanh}(\mathbf{z}_{cover}) / \kappa$\;
    $\mathbf{c}_{est} \gets \text{clamp}(\mathbf{c}_{est}, -1.0, 1.0)$\;
}
\Return{$\mathbf{c}_{est}$}\;
\end{algorithm}

\vspace{-0.5em}
\subsection{Removal Architecture}
\vspace{-0.5em}
In addition to watermark embedding, FiLM-CAM supports watermark removal without introducing a separate removal network. Instead, we reuse the {\em same embedder} $E$ as a residual estimator at removal time. 
Let $x_{w}$ denote a watermarked image generated from cover image $x$, secret $m$, and key $k$. During removal, the embedder is applied again to the watermarked image to predict a removal residual $r_{\mathrm{rem}} = E(x_{w}, m, k)$. Removal therefore takes three inputs: the watermarked image $x_w$, the key $k$, and the embedded message $m$. In practice, an authorized party holding $k$ can first recover $m$ by decoding $x_w$ with the extractor, so the key is the only secret required. Supplying the correct message improves residual estimation and removal fidelity, while with a wrong key removal fails regardless of the message (Appendix~\ref{app:removal_inputs}).
This residual is then used to invert the watermark composition and recover an estimate of the underlying cover image; $\hat{x} = C^{-1}(x_{w}, r_{\mathrm{rem}}),$
where $C^{-1}(\cdot)$ denotes the inverse of the composition operator (\Cref{sec:watermark_composition}).

As FiLM-CAM composes the watermark in transformed logit space rather than by direct residual addition, watermark removal must follow the corresponding inverse transformation. In the forward process, the image-dependent gain is computed from the clean cover image. In practice, during removal, the cover image may not be available to the user, so the true gain cannot be evaluated directly. We therefore approximate the inverse using a small number of iterative refinement steps (\Cref{fig:rewatermark_alpha_refine}, bottom right) that mirror the forward composition, recomputing the gain from the current cover estimate. The full procedure is given in \cref{alg:invert_watermark}. Starting from the watermarked image as an initial estimate, the algorithm repeatedly updates the gain using the current estimate, removes the predicted transformed-space perturbation, and maps results to the valid image range.

\vspace{-2mm}
\subsection{Training Schedule}
\label{sec:sched}
\vspace{-2mm}

We train FiLM-CAM in two stages. Stage 1 learns key-conditioned watermark embedding, while Stage 2 fine-tunes the embedder to be effective for watermark removal as well. This separation is useful because Stage 1 must balance image fidelity and message recoverability for an initially unwatermarked input, whereas Stage 2 must add capability to invert the embedding from a watermarked input.

{\textbf{Stage 1: Key-conditioned watermark embedding.}}
Given a cover image $x$, secret message $m$, and key $k$, the embedder predicts a residual $r = E(x,m,k)$,
which is composed with the cover image to obtain the watermarked image $x_{w} = C(x,r)$.  After the noise layer, the extractor predicts the embedded message as: $\hat{m} = X(\tilde{x}_{w},k), \ \ \tilde{x}_{w} = N(x_{w}).$ The Stage 1 objective follows standard formulations in neural watermarking and balances image fidelity against message recovery. For the correct key, the total loss is
\begin{eqnarray}
L_{\mathrm{S1}}^{\mathrm{corr}} =
\lambda_{\mathrm{img}} L_{\mathrm{img}}(x,x_{w})
+
\lambda_{\mathrm{sec}} L_{\mathrm{BCE}}(\hat{m},m)
+
\lambda_{\mathrm{adv}} L_{\mathrm{GAN}},
\end{eqnarray}
where $L_{\mathrm{img}}$ is the image-quality loss, encouraging high‑quality watermarked images, $L_{\mathrm{BCE}}$ is the binary cross-entropy message-recovery loss, encouraging accurate decoding, and $L_{\mathrm{GAN}}$ is an adversarial generator objective. Following \cite{bui2025trustmark}, the image-quality term combines spatial, perceptual, and frequency-domain fidelity terms:
\begin{eqnarray}
L_{\mathrm{img}}(x,x_{w})
=
\lambda_{\mathrm{YUV}} L_{\mathrm{YUV}}
+
\lambda_{\mathrm{LPIPS}} L_{\mathrm{LPIPS}}
+
\lambda_{\mathrm{FFL}} L_{\mathrm{FFL}},
\end{eqnarray}
where $L_{\mathrm{YUV}}$ is the reconstruction loss in YUV space, $L_{\mathrm{LPIPS}}$ is the perceptual loss, and $L_{\mathrm{FFL}}$ is the focal frequency loss. In addition, we train the embedder in a generative adversarial manner using the adversarial setup as in TrustMark \cite{bui2025trustmark}. To enforce conditional access, we additionally train the model under wrong-key conditions. Let $k' \neq k$ denote an incorrect key, and let $\hat{m}' = X(\tilde{x}_{w},k')$ be the extracted message under the wrong key. In this case, we keep the same image quality and adversarial objectives, but replace the recovery target with an invalid decoding target. We use the all-zero vector $0$ for convenience, but the choice of zeros is arbitrary, and we find it does not affect performance.
\begin{eqnarray}
L_{\mathrm{S1}}^{\mathrm{wrong}} =
\lambda_{\mathrm{img}} L_{\mathrm{img}}(x,x_{w})
+
\lambda_{\mathrm{sec}} L_{\mathrm{BCE}}(\hat{m}',0)
+
\lambda_{\mathrm{adv}} L_{\mathrm{GAN}}.
\end{eqnarray}
Wrong‑key training encourages visually similar watermarked images ($L_{\mathrm{img}}$ and $L_{\mathrm{GAN}}$), and a decoding collapse to an invalid codeword (via $L_{\mathrm{BCE}}$ with the all‑zero target). Each training iteration contains one correct-key step followed by one wrong-key step for the same embedded sample. In the correct-key step, the embedder and extractor are trained so that the watermark remains recoverable under the authorized key. In the wrong-key step, the same embedding pipeline is retained, but the extractor is evaluated with an incorrect key and trained toward the invalid target. This encourages a gap between authorized and unauthorized decoding. Stage 1 uses a training strategy in which the model first prioritizes message recovery, after which stronger augmentations and full image fidelity pressure are progressively introduced. In particular, the noise layer is activated gradually, and the image-loss weighting is ramped during training rather than applied at full strength from the start.

{\textbf{Stage 2: Key-conditioned watermark removal.}}
In Stage 2, a trainable embedder learns to remove the watermark. A frozen copy of the Stage 1 embedder is first used to generate the watermark residual and to create the watermarked image, while the trainable embedder (from last stage 1) is then trained on that watermarked image to invert the embedding process. Formally, given a clean image $x$, secret $m$, and key $k$, the frozen Stage 1 embedder produces a reference residual $r^\star = E_{\mathrm{frozen}}(x,m,k)$, which is composed with the cover image to obtain a watermarked image $x_{w} = C(x,r^\star)$. The trainable embedder is then applied to the watermarked image to predict a removal residual
$r_{\mathrm{rem}} = E(x_{w},m,k)$.  
As the watermark composition is performed in transformed logit space, we train removal in the corresponding transformed perturbation space. Using the gain term defined in \Cref{sec:watermark_composition}, we compute: $\Delta z^\star = \alpha \tanh(r^\star) \odot g(x_w), \ \
\Delta z_{\mathrm{rem}} = \alpha \tanh(r_{\mathrm{rem}}) \odot g(x_w),$ and define the correct-key removal objective, removing the residual matches embedded in the transformed space as an $L_1$ consistency loss: $L_{\mathrm{S2}}^{\mathrm{corr}} = \|\Delta z_{\mathrm{rem}} - \Delta z^\star\|_1$. 
For a wrong key $k' \neq k$, the predicted residual should not correspond to the true watermark signal; we therefore encourage the transformed perturbation to vanish: $r_{\mathrm{rem}}' = E(x_{w},m,k')$;  $\Delta z_{\mathrm{rem}}' = \alpha \tanh(r_{\mathrm{rem}}') \odot g(x_{w})$; $L_{\mathrm{S2}}^{\mathrm{wrong}} = \left\|\Delta z_{\mathrm{rem}}'\right\|_1$.
Thus, under the correct key, the trainable embedder learns to recover the true embedding perturbation, whereas under a wrong key, it is attenuated.

We first train Stage 1 for 120 epochs, then alternate between Stage 1 and Stage 2 in 5-epoch cycles. This alternating scheme allows the model to refresh watermark embedding quality while progressively improving key-conditioned removal behavior. In practice, wrong-key removal only partially suppresses the watermark (post-removal bit accuracy remains well above chance), and reconstruction quality is worse than under the correct key (cf. \Cref{sec:eval_wmremoval}). FiLM and the noise layer play complementary roles. FiLM is the source of conditionality: it makes the predicted residual key-dependent, so a wrong key yields a different residual (\Cref{fig:residual_comparison}). An extractor trained without noise layer is sensitive to any perturbation subtracted from the watermarked image, so even a wrong-key residual collapses decoding. Training with the noise layer makes the extractor robust to such perturbations.

\vspace{-2mm}
\section{Experiments and Discussion}
\label{sec:eval}
\vspace{-2mm}
We evaluate FiLM-CAM in terms of visual quality and robustness, and its conditional-access properties, resistance to decoding under incorrect keys and the quality of authorized watermark removal.

\vspace{-0.5em}
\subsection{Datasets, metrics and training details}
\vspace{-0.5em}



\textbf{Datasets.} We train FiLM-CAM on the MIRFlickr-1M\cite{huiskes2008mir} dataset. Specifically, we use 100K images for training and 1K images for validation. For evaluation, we test on three datasets: COCO\cite{lin2014microsoft} (5K images), DIV2K\cite{agustsson2017ntire} (100 images), and SA-1B\cite{kirillov2023segment} (5K images). COCO provides large-scale natural image diversity, DIV2K contains high-quality images with higher spatial fidelity, and SA-1B introduces additional real-world variability at scale. Unless otherwise stated, each test image is assigned a random binary watermark and a randomly sampled key.
\vspace{-0.5em}

\noindent\textbf{Metrics.} We measure imperceptibility with PSNR, SSIM and LPIPS~\cite{zhang2018unreasonable}, and decoding with the bit accuracy (\BA) of the recovered message, where 0.5 is random guessing; BER $=1-$\BA (\%). \PSNRW and \PSNRR denote the PSNR of the watermarked and the watermark-removed image w.r.t.\ the original cover, \dPSNR $=$ \PSNRR $-$ \PSNRW, and \BAR is the \BA after removal. $\checkmark$/$\times$ denote the correct/wrong key, and $\uparrow$/$\downarrow$ mark higher/lower is better. With a wrong key the desired behaviour is reversed: \BA should approach 0.5, and \BAR should stay high so the watermark survives. Unless otherwise stated, all methods are evaluated at the original image resolution.
\vspace{-0.5em}

\begin{table*}[t!]
  \caption{Comparison with SOTA neural watermarking on COCO, DIV2K and SA-1B.}
  \label{tab:sota_three_datasets}
  \centering
  \setlength{\tabcolsep}{2.0pt}
  \scriptsize
  \begin{tabular}{@{}lc|cccc|cccc|cccc@{}}
    \toprule
   & & \multicolumn{4}{c|}{COCO} & \multicolumn{4}{c|}{DIV2K} & \multicolumn{4}{c}{SA-1B} \\
    Method & Payload & \BA & \PSNRW & SSIM & LPIPS
           & \BA & \PSNRW & SSIM & LPIPS
           & \BA & \PSNRW & SSIM & LPIPS \\
    \midrule
    TrustMark-Q \cite{bui2025trustmark} & 100
      & 0.99 & 41.9 & 0.9910 & 0.0017
      & 0.99 & 40.9 & 0.9939 & 0.0026
      & 0.99 & 41.3 & 0.9946 & 0.0028 \\
    InvisMark \cite{xu2024invismark} & 100
      & 0.99 & 48.0 & 0.9974 & 0.0011
      & 0.99 & 46.5 & 0.9964 & 0.0025
      & 0.99 & 46.2 & 0.9961 & 0.0027 \\
    PixelSeal \cite{souvcek2025pixel} & 256
      & 0.99 & 45.8 & 0.9948 & 0.0020
      & 0.99 & 46.0 & 0.9959 & \textbf{0.0011}
      & 0.99 & 46.5 & 0.9966 & \textbf{0.0012} \\
    RoSteALS \cite{bui2023rosteals} & 100
      & 0.99 & 30.0 & 0.9462 & 0.0475
      & 0.98 & 28.2 & 0.9484 & 0.0527
      & 0.99 & 28.6 & 0.9537 & 0.0521 \\
    SSL \cite{fernandez2022watermarking} & 100
      & 0.99 & 40.0 & 0.9788 & 0.0351
      & 0.99 & 40.0 & 0.9733 & 0.0611
      & 0.99 & 40.0 & 0.9770 & 0.0350 \\
    MBRS \cite{jia2021mbrs} & 256
      & 0.99 & 43.2 & 0.9729 & 0.0041
      & 0.99 & 42.2 & 0.9774 & 0.0043
      & 0.99 & 42.5 & 0.9755 & 0.0047 \\
    Key-based FNNS \cite{luo2023securing} & 1\,bpp
      & 0.99 & 40.4 & 0.9800 & -
      & 0.99 & 33.9 & 0.9600 & -
      & -    & -    & - & - \\
    \midrule
    \textbf{FiLM-CAM} ($\checkmark$ key) & 100$^{\dagger}$
      & \textbf{0.99} & \textbf{48.2} & \textbf{0.9986} & \textbf{0.0010}
      & \textbf{0.99} & \textbf{46.7} & \textbf{0.9975} & 0.0020
      & \textbf{0.99} & \textbf{46.6} & \textbf{0.9974} & 0.0020 \\
    \textbf{FiLM-CAM} ($\times$ key) & 100$^{\dagger}$
      & \textbf{0.53} & - & - & -
      & \textbf{0.54} & - & - & -
      & \textbf{0.53} & - & - & - \\
    \midrule
    \textbf{FiLM-CAM (noise)} ($\checkmark$ key) & 100$^{\dagger}$
      & \textbf{0.99} & \textbf{40.2} & 0.9959 & 0.0031
      & \textbf{0.99} & \textbf{40.1} & 0.9953 & 0.0054
      & \textbf{0.99} & \textbf{40.1} & 0.9953 & 0.0053 \\
    \textbf{FiLM-CAM (noise)} ($\times$ key) & 100$^{\dagger}$  
      & \textbf{0.55} & - & - & -
      & \textbf{0.55} & - & - & -
      & \textbf{0.55} & - & - & - \\
    \bottomrule
  \end{tabular}
  \vspace{2pt}
\parbox{\textwidth}{\scriptsize $^{\dagger}$FiLM-CAM embeds a 100-bit watermark payload plus a separate 100-bit secret key (200 bits of information in total); only the 100-bit payload is recovered at decoding, and the key is not counted in capacity. bpp: bits per pixel}

\end{table*}

\noindent\textbf{Training details.} Our FiLM-CAM configuration uses a 100-bit watermark payload ($n$) and a 100-bit binary key ($L$), with the key mapped to a 64-dimensional embedding ($d$) shared between the embedder and extractor. We train with a batch size of $32$ using AdamW with a base learning rate of $4 \times 10^{-5}$ and a cosine-annealed schedule. In Stage 1, we use reconstruction weight $1.5$, perceptual weight $20.0$, secret recovery weight $40.0$, focal frequency loss weight $1.0$, generator weight $10$, and discriminator weight $20$. The image-loss ramp is increased linearly from $0$ to $100$ over $90$K steps, starting $ 50$K steps after the delay. For logit-space watermark composition, we use $\kappa = 0.95$, $\epsilon = 10^{-3}$, and watermark strength $\alpha = 100$. No fixed key set is used: keys are freshly sampled uniformly at random from $\{0,1\}^{100}$ at every iteration, and wrong keys are sampled independently. Since the key space is $2^{100}$, test-time keys are unseen by construction, and correct-key extraction at test time directly measures generalization to new keys. For conditional training, we use one wrong-key step per correct-key step, after a key warm-up of $10$ epochs, during which only correct keys are used. Stage 2 removal training is enabled after $120$ initial Stage 1 epochs, after which we alternate between Stage 1 and Stage 2 every $5$ epochs. The Stage 2 consistency loss is $L_1$ with $40.0$ weight. Unless stated otherwise, we report results for FiLM-CAM trained under clean conditions for imperceptible watermarking.
\vspace{-0.5em}
\subsection{Comparison with the state-of-the-art}
\vspace{-0.5em}
We compare FiLM-CAM against state-of-the-art (SOTA) neural watermarking baselines: TrustMark~\cite{bui2025trustmark}, InvisMark~\cite{xu2024invismark}, PixelSeal~\cite{souvcek2025pixel}, RoSteALS~\cite{bui2023rosteals}, SSL~\cite{fernandez2022watermarking}, and MBRS~\cite{jia2021mbrs}. Although no prior end-to-end deep image watermarking method explicitly supports key-conditioned CAM, SSL~\cite{fernandez2022watermarking} is arguably the closest comparison, as its carrier-based design introduces per-bit embedding vectors that could be repurposed to emulate key-like behavior. In addition, we compare against a key-based FNNS\cite{luo2023securing} method in our conditional-access experiments.
\vspace{-0.5em}

\begin{figure}[t]
  \centering
  \includegraphics[width=1\linewidth]{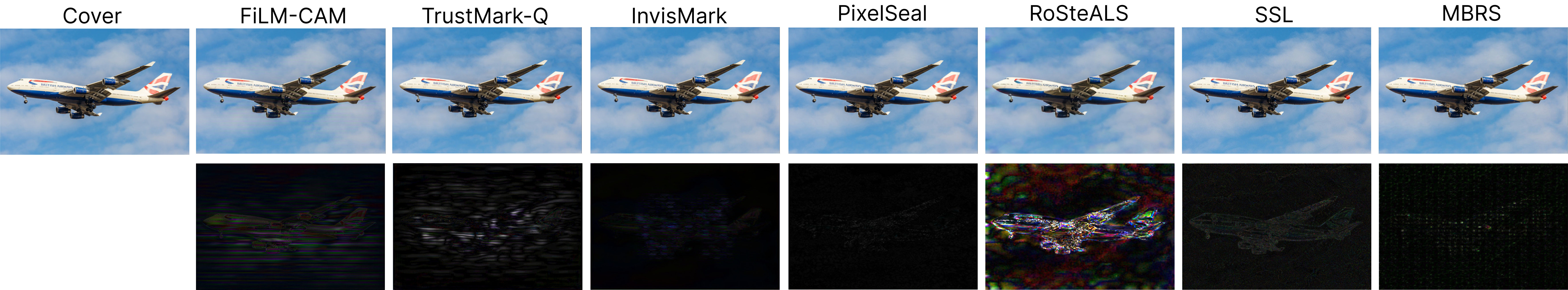}
  \caption{Watermarked images (top) and amplified absolute residuals $|x_w - x|$ (bottom) for FiLM-CAM and baselines on the same cover.}

  \label{fig:qualitative_comparison_all_methods}
\end{figure}

\textbf{Watermark Embedding.}
We first compare FiLM-CAM against baselines in the standard watermark embedding setting. \Cref{tab:sota_three_datasets} reports watermark extraction bit accuracy and watermarked-image PSNR across COCO, DIV2K, and SA-1B. FiLM-CAM is decoded using the correct key (FiLM-CAM ($\checkmark$ key)). The results show that FiLM-CAM is competitive with SOTA, achieving near-perfect decoding $\sim$0.99 bit accuracy across the datasets while maintaining a PSNR of $\sim$46 dB. Qualitative comparisons of watermarked images and their amplified residuals against the baselines are shown in \Cref{fig:qualitative_comparison_all_methods}.  We train a robustness-oriented variant of FiLM-CAM with the noise layer, sampling a mixture of common geometric, photometric perturbations, and compression artifacts. On DIV2K, we attain a mean bit-accuracy of $\sim$0.96 across all these attacks. Appendix~\ref{app:robustness} for full  per-transform results. The noise-trained variant has lower embedding PSNR ($\sim$40 vs.\ $\sim$47\,dB); this is the cost of a robust extractor, which is also what makes unauthorized removal fail.
\vspace{-0.5em}

\begin{figure}[t]
  \centering
  \includegraphics[width=0.65\linewidth]{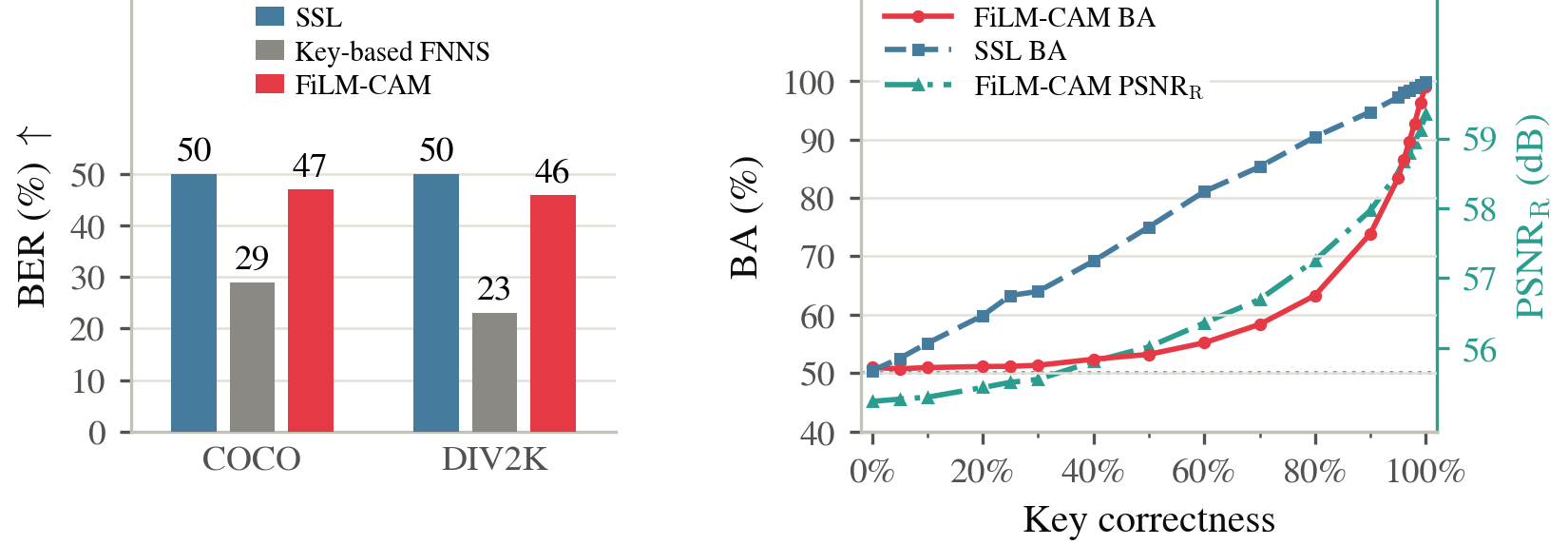}
  \caption{(Left) Wrong-key BER on COCO and DIV2K. (Right) \BA of FiLM-CAM and SSL, and \PSNRR of FiLM-CAM, vs.\ the fraction of correct key bits.}
  \label{fig:key_correctness}
\end{figure}

\noindent\textbf{Conditional Watermarking}
We next evaluate the conditional-access behavior of FiLM-CAM. We show the desired asymmetric behavior: when an image is watermarked with a key $k_A$, the embedding message is recoverable only when decoding is performed with the same key $k_A$, whereas decoding with any incorrect key $k_B\neq k_A$ fails. 
To quantify this behavior, we first compare FiLM-CAM against SSL \cite{fernandez2022watermarking} and a key-based FNNS \cite{luo2023securing} method using bit error rate (BER) under wrong-key or wrong-carrier(for SSL) decoding. Here, a higher BER is better, since it indicates a stronger failure under unauthorized decoding. The results are summarized in \Cref{fig:key_correctness} (left) and indicate that FiLM-CAM provides stronger wrong-key suppression than the FNNS baseline, i.e., approaching the ideal behavior of random decoding.

We further analyze conditional access by varying the matching between the decoding key (query key) and the true embedding key. We define 'key-correctness' as the percentage of bits in a query key $k_B$ that match the true embedding key $k_A$.  \Cref{fig:key_correctness} (right) plots bit accuracy (using $k_B$) as a function of key-correctness for FiLM-CAM and SSL. For FiLM-CAM, bit accuracy remains close to random guessing over a wide range of key-correctness \%, and rises sharply only when query key $k_B$ becomes very close to the true embedding key $k_A$. Compared to SSL, FiLM-CAM is more key-specific sustaining around random (50-60\%) bit-accuracy up to roughly 70\% key-correctness, and only approaches high bit-accuracies ($\sim$95\%) when nearly all the key is correct ($\sim$98\%).

\vspace{-2.5mm}
\subsection{Watermark Removal}
\label{sec:eval_wmremoval}
\vspace{-0.5em}

We compare against several learning and regeneration-based attacks. 
We include \textit{WatermarkAttacker}~\cite{zhao2024invisible},
a regeneration-based attack framework, and follow its implementation to evaluate three instantiations: VAE-Bmshj2018\cite{balle2018variational} and VAE-Cheng2020\cite{cheng2020learned} and a diffusion-based \cite{rombach2022high} regeneration variant (Diffusion Attack) as in \cite{zhao2024invisible}. We also include Ctrl-Regen+\cite{liu2024ctrlregen}, a controlled image regeneration framework designed to erase watermarks while maintaining semantic fidelity. 
 Finally, we report results for TrustMark-RM (ReMark)~\cite{bui2025trustmark}, a removal model introduced in TrustMark.


\begin{table*}[t!]

    \caption{Watermark removal on COCO, DIV2K and SA-1B, comparing removal attacks against key-conditioned removal with the noise-trained FiLM-CAM. \dPSNR is measured relative to the watermarked image, and time is per image. \BAR should approach 0.5 (watermark erased) for the baselines and for FiLM-CAM with the correct key, but should stay high for FiLM-CAM with the wrong key, where the watermark must survive the unauthorized removal attempt; this column therefore has no arrow.}
  \label{tab:removal_comparison}
  \centering
  \scalebox{0.94}{
  \scriptsize
  \setlength{\tabcolsep}{3.5pt}
  \begin{tabular}{@{}lccc|ccc|ccc|c@{}}
    \toprule
    & \multicolumn{3}{c|}{COCO} & \multicolumn{3}{c|}{DIV2K} & \multicolumn{3}{c|}{SA-1B} & \multicolumn{1}{c}{Inference}\\
    Method
    & \PSNRR$\uparrow$ & \BAR & \dPSNR$\uparrow$
    & \PSNRR$\uparrow$ & \BAR & \dPSNR$\uparrow$
    & \PSNRR$\uparrow$ & \BAR & \dPSNR$\uparrow$ & time (ms)$\downarrow$\\
    \midrule
    TrustMark-RM(Q) \cite{bui2025trustmark}      & 41.7 & 0.66 & -0.2 & 40.8 & 0.74 & -0.2 & 41.08 & 0.72 & -0.2&35.9\\
    CtrlRegen+ \cite{liu2024ctrlregen}           & 21.1 & 0.50  & -27.1 & 20.0 & 0.51 & -26.7 & 20.0 & 0.50 & -26.6&1976 \\
    Diffusion Attack \cite{zhao2024invisible}            &21.0  & 0.50 & -27.2 & 22.4 & 0.51 & -24.3 & 22.0 & 0.50 & -24.6&255.63  \\
    VAE-Bmshj2018 \cite{balle2018variational}           & 30.0 & 0.50 & -18.2 & 25.3 & 0.51 & -21.4 & 25.5 & 0.50 & -21.1&4.00 \\
    VAE-Cheng2020 \cite{cheng2020learned}           & 31.4 & 0.50 & -16.8 & 26.4 & 0.51 &-20.3  & 26.6 & 0.50 & -20.0 &14.46 \\
    \midrule
    \textbf{FiLM-CAM (noise) ($\checkmark$ key)} & \textbf{57.9} & 0.57 & \textbf{17.7} & \textbf{54.0} & 0.58 & \textbf{13.9} & \textbf{54.0} & 0.58 & \textbf{13.9}&5.06 \\
    \textbf{FiLM-CAM (noise) ($\times$ key)} & 48.0 & \textbf{0.81} & 7.8 & 46.6 & \textbf{0.79} & 6.5 & 46.7 & \textbf{0.78} & 6.6 &5.06\\
    \bottomrule 
  \end{tabular}
  }
\end{table*}

\noindent\textbf{Removal Performance.} \Cref{tab:removal_comparison} reports removal PSNR, post-removal bit-accuracy, and
PSNR gain relative to the watermarked image. Under the correct key, FiLM-CAM
attains very high removal PSNR ($\sim\!58$\,dB on COCO and $\sim\!54$\,dB on
DIV2K and SA-1B), while post-removal bit-accuracy drops close to random
($\sim\!0.57$--$0.58$), showing that the logit-space inversion removes the
embedded signal without sacrificing fine image details. Most regeneration-based
baselines achieve near-random post-removal bit-accuracy but at substantially
lower PSNR ($\sim\!20$--$30$\,dB); TrustMark-RM \cite{bui2025trustmark} preserves
visual quality (gain $\sim\!-0.2$\,dB) but does not fully suppress decoding,
with post-removal bit-accuracies well above random. Crucially, FiLM-CAM
exhibits a sharp asymmetry between authorised and unauthorised removal: with
the wrong key, reconstruction quality drops to $\sim\!47$\,dB \emph{and}
post-removal bit-accuracy remains high ($\sim\!0.78$--$0.81$), so the watermark
survives the removal attempt and stays decodable. 
This conditional behaviour
is enforced by training the model with a noise layer; without it, removal
generalises across keys and the watermark can be destroyed regardless of key
correctness (see \Cref{tab:filmcam_clean_vs_noise_appexdix} in Appendix).

\noindent{{\textbf{Iterative Refinement for Removal.}}}
We analyze the effect of iterative refinement (\Cref{fig:rewatermark_alpha_refine}, bottom right). 
With the correct key, the PSNR of removal consistently improves as the number of refinement steps increases, eventually surpassing the cover-guided baseline. In contrast, removal with a wrong key quickly saturates at a lower PSNR, close to the wrong-key cover-guided baseline. 
Diminishing returns exist >$20$ iterations, motivating our choice of 20 refinement steps as a practical trade-off between quality and computational cost.

\vspace{-2mm}
\subsection{Re-Watermarking} \label{sec:re_watermarking}
\vspace{-2mm}

\paragraph{\textbf{{Repeated Watermark Embedding }\textbf{(RWME):}}} Starting from a clean image $x_0$, we repeatedly apply the embedder, sampling a fresh (message, key) pair at each iteration: $x^{\mathrm{w}}_{t+1} \leftarrow \mathrm{Embed}\!\left(x^{\mathrm{w}}_{t}\right), \ \ t = 0,\ldots,T-1,$
where $x^{\mathrm{w}}_{0} = \mathrm{Embed}(x_0)$. We report the watermark bit-accuracy and the PSNR of $x^{\mathrm{w}}_{t}$ relative to the original cover $x_0$ as a function of the iteration index $t$.
\paragraph{\textbf{{Embed--Remove Cycling (ERC):}}}
Starting from a clean image $x_0$, we alternate between embedding and authorized removal:
$x^{\mathrm{w}}_{t} \leftarrow \mathrm{Embed}(x_t)$;\;  and \
$x_{t+1} \leftarrow \mathrm{Remove}(x^{\mathrm{w}}_{t})$,\; for $t = 0,\ldots,T-1$.

We report (i) the PSNR of the watermarked images $x^{\mathrm{w}}_{t}$ relative to the original $x_0$, (ii) the PSNR of the reconstructed images $x_t$ relative to $x_0$, and (iii) bit-accuracies at each cycle.

\begin{figure}[t!]
  \centering
  \includegraphics[width=0.75\linewidth]{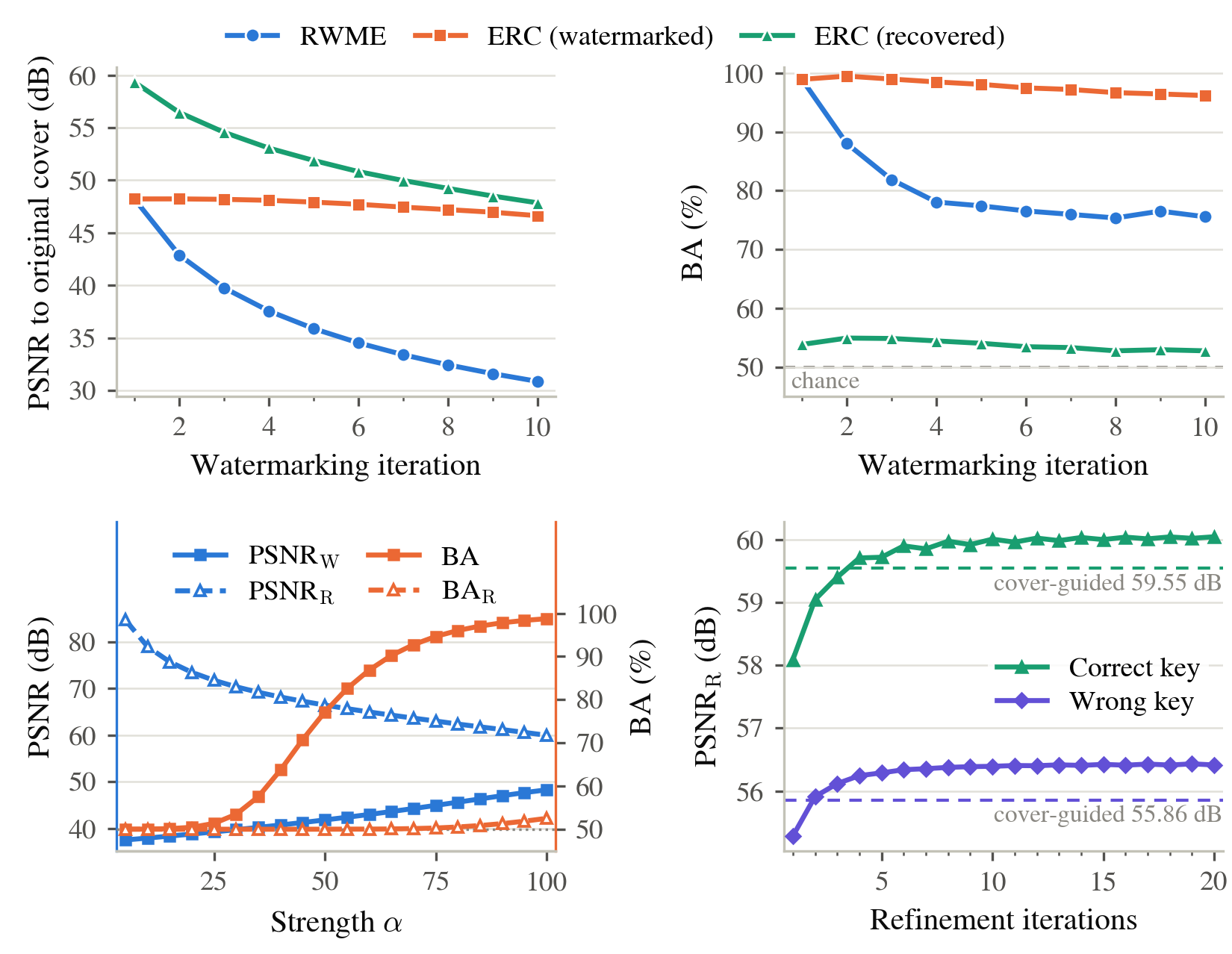}
  \vspace{-0.6em}
  \caption{\textbf{Top:} re-watermarking, comparing RWME (blue) with ERC (orange: watermarked image, green: recovered image); (left) PSNR w.r.t.\ the original cover, (right) \BA. \textbf{Bottom left:} effect of watermark strength $\alpha$ on \PSNRW and \PSNRR (blue, left axis) and on \BA and \BAR (orange, right axis). \textbf{Bottom right:} \PSNRR vs.\ number of refinement iterations; dashed lines show removal using the true cover's gain.}
  \label{fig:rewatermark_alpha_refine}
\end{figure}

\noindent \Cref{fig:rewatermark_alpha_refine} (top) summarizes the results. In RWME, repeated embedding accumulates distortion across iterations, leading to a steep drop in PSNR and bit-accuracy. In ERC, the watermarked-image PSNR remains steady across cycles, whereas the reconstructed-image PSNR decreases gradually, indicating that each embedding-removal step incurs a small reconstruction error that accumulates over cycles.

\vspace{-2.5mm}
\subsection{Ablation studies}
\label{sec:ablation}
\vspace{-2.5mm}

\begin{table}[t]
    \caption{Ablations of FiLM-CAM (clean) on DIV2K with the correct key. Each row changes one design choice and keeps all other settings fixed; best values in bold.}
  \label{tab:ablations}
  \centering
  \scriptsize
  \setlength{\tabcolsep}{3pt}
  \begin{subtable}[t]{0.42\linewidth}
    \centering
    \caption{FiLM-CAM components.}
    \label{tab:ablation_filmcam}
    \setlength{\tabcolsep}{2pt}
    \begin{tabular}{@{}lcccc@{}}
      \toprule
      Method & \BA$\uparrow$ & \PSNRW$\uparrow$ & \PSNRR$\uparrow$ & \BAR$\downarrow$ \\
      \midrule
      w/o FiLM       & 0.85 & 36.9 & 47.3 & 0.62 \\
      w/o gain       & 0.99 & 36.0 & 44.0 & \textbf{0.51} \\
      Additive comp. & 0.98 & \textbf{47.0} & 54.1 & 0.55 \\
      \midrule
      \textbf{FiLM-CAM} & \textbf{0.99} & 46.7 & \textbf{55.4} & 0.54 \\
      \bottomrule
    \end{tabular}
  \end{subtable}\hfill
  \begin{subtable}[t]{0.56\linewidth}
    \centering
    \caption{Encoder--decoder backbones.}
    \label{tab:ablation_backbones}
    \begin{tabular}{@{}llcccc@{}}
      \toprule
      Encoder & Decoder & \BA$\uparrow$ & \PSNRW$\uparrow$ & \PSNRR$\uparrow$ & \BAR$\downarrow$ \\
      \midrule
      MUNIT~\cite{bui2025trustmark} & ResNet-50~\cite{he2016deep}  & 0.97 & 41.3 & 54.0 & 0.67 \\
      Ours & ResNet-50~\cite{he2016deep}                            & 0.96 & 37.5 & 43.1 & 0.67 \\
      Ours & ConvNeXt-T~\cite{liu2022convnet}                       & 0.99 & 38.6 & 49.8 & 0.65 \\
      Ours & ConvNeXtV2-B~\cite{woo2023convnext}                    & 0.98 & 38.2 & 50.5 & 0.60 \\
      \midrule
      Ours & ConvNeXt-B~\cite{liu2022convnet} & \textbf{0.99} & \textbf{46.7} & \textbf{55.4} & \textbf{0.54} \\
      \bottomrule
    \end{tabular}
  \end{subtable}
\end{table}


\vspace{0.3mm}
\noindent\textbf{Effect of $\alpha$ (Watermark Strength).}
We study the influence of parameter $\alpha$ (\Cref{fig:rewatermark_alpha_refine}, bottom left). Increasing $\alpha$ strengthens the watermarking signal, which trades off robustness (bit-accuracy) against reconstruction imperceptibility (PSNR).

\noindent\textbf{FiLM-CAM components.} We conduct a series of ablations to justify our design choices (\Cref{tab:ablation_filmcam}). \emph{Removing FiLM conditioning:} we replace FiLM conditioning with direct key input to the embedder. Without FiLM, training does not converge to the high decoding accuracy of the full model: \BA drops (0.85 vs.\ 0.99), and removal quality degrades (\PSNRR 47.3\,dB vs.\ 55.4\,dB). \emph{Effect of the gain factor:} we remove the gain factor from the embedding formulation. While \BA remains high, both \PSNRW and \PSNRR decrease substantially (\PSNRR 44.0\,dB vs.\ 55.4\,dB). Without the gain, we also cannot perform iterative refinement to progressively improve \PSNRR. \emph{Logit-space vs.\ additive composition:} we replace logit-space composition with plain additive composition ($x_w = x + r$), keeping all else fixed. Both variants give comparable embedding quality; the benefit of logit-space composition is in removal: \PSNRR improves by 1.3\,dB (55.4 vs.\ 54.1\,dB), i.e.\ the watermark is erased more completely and the cover is restored more faithfully. On COCO, the gap is larger: 59.3 vs.\ 55.5\,dB \PSNRR and 0.52 vs.\ 0.56 \BAR.

\vspace{-\parskip}
\noindent{{\textbf{Encoder--Decoder Backbone Choices.}}}
Finally, we evaluate different encoder--decoder backbones (\Cref{tab:ablation_backbones}). 
In addition to our encoder (Ours), we consider a MUNIT-based encoder \cite{bui2025trustmark} modified to include CAM and logit-space embedding. Our encoder paired with a ConvNeXt-Base extractor provides the strongest performance, but the results also show that CAM is effective across several encoder--decoder backbones.

\vspace{-2mm}
\section{Conclusion}
\vspace{-2mm}

We presented FiLM-CAM, a conditional-access deep watermarking method that enables key-dependent watermark embedding, decoding, and removal within a single encoder–decoder architecture. By integrating a lightweight Conditional Access Module (CAM) based on feature-wise linear modulation (FiLM), FiLM-CAM conditions internal network representations on a secret key, allowing a shared watermarking backbone to support multiple private watermarking domains. Our logit-space watermark composition formulation enables watermark removal by estimating the residual using the same embedder network, trained to predict residual invariance in the presence of a watermark. We showed that FiLM-CAM performs competitively in terms of quality and accuracy, whilst introducing conditional access functionality that denies reliable decoding under incorrect keys. 

\vspace{-\parskip}
\noindent\textbf{Limitations.} Conditional access prevents key-free \emph{decoding} and key-free \emph{reconstructive} removal, but not key-free \emph{destruction}. Re-embedding with another key (Appendix~\ref{app:overwrite}) or strong diffusion-based regeneration (Appendix~\ref{app:waves}) disables detection without the key. This holds for learned watermarking methods generally, but such attacks do not recover the original cover image.

\vspace{-\parskip}
\noindent \textbf{Future Work} could explore raising bit capacity or whether additional side-information could be created during watermark encoding that could be retained as a `hint' to further raise removal image quality.  Both wrong-key decoding and
wrong-key removal are trained to fail by design --- decoding collapses to
random, and removal is intended to preserve rather than destroy the watermark.
While a substantial portion of the watermark survives an unauthorised removal
attempt, the full payload is not always retained; a stronger objective that
pushes post-removal bit-accuracy under wrong keys closer to the originally
embedded message would further sharpen this authorised/unauthorised asymmetry.

\begin{ack}

This research was funded in part by Adobe Research and in part by DECaDE under the UKRI/EPSRC Grant reference EP/T022485/1.  Attendance of Yash Kulthe to present at NeurIPS 2026 was supported in part by NeurIPS through a financial assistance grant.%

\end{ack}

{
\small
\bibliographystyle{plainnat}  
\bibliography{main}
}

\clearpage

\section*{Appendix}


\appendix

\section{Robustness Results}
\label{app:robustness}
\FloatBarrier  
Our noise layer samples a mixture of common geometric and photometric perturbations, along with compression artifacts, to improve decoding robustness to image distortions.  In addition to the FiLM-CAM results in the main paper, we present an additional robustness variant using a ConvNeXtV2-Tiny\cite{woo2023convnext} extractor backbone. Table~\ref{tab:noise_layer} summarizes the training-time noise layer used to train the model. Concretely, after watermark composition, we apply a fixed set of transforms, along with two additional transforms sampled uniformly from a pool that includes JPEG compression, photometric adjustments, blur, additive noise, posterization, RGB shift, and geometric distortions such as perspective change, resizing, and random cropping. The augmentation strength is gradually increased during training using a ramp schedule.

We report a detailed robustness comparison against prior watermarking methods in Table~\ref{tab:robustness_div2k2}. Across the perturbation suite on DIV2K\cite{agustsson2017ntire}, the robustness-trained FiLM-CAM model attains a mean extraction bit accuracy of $96.0\%$. This shows that FiLM-CAM remains highly decodable under a broad range of common image corruptions and transformations, while retaining the conditional-access design introduced in the main paper. We additionally evaluate the noise-trained model on the standardized WAVES benchmark \cite{an2024waves}, reporting TPR@0.1\%FPR across distortion, regeneration, rinsing and adversarial attack families (Appendix~\ref{app:waves}).

Finally, Table~\ref{tab:filmcam_clean_vs_noise_appexdix} compares the clean and robustness-trained FiLM-CAM models across COCO\cite{lin2014microsoft}, DIV2K\cite{agustsson2017ntire}, and SA-1B\cite{kirillov2023segment}. As expected, introducing the noise layer leads to a modest drop in clean-condition image quality and some metrics related to embedding and removal performance. However, the robustness-trained model still preserves strong correct-key extraction, wrong-key decoding remains near chance, and post-removal behavior stays broadly consistent with the clean model. Overall, this reflects the expected trade-off: the noise-trained variant gives improved robustness to perturbations at the cost of a minor decrease in clean-condition scores.

\begin{table*}[t]
\centering
\scriptsize
\setlength{\tabcolsep}{4pt}
\caption{Training-time noise layer used in FiLM-CAM, as implemented in our augmentation code. Two optional transforms are sampled uniformly at each forward pass (\texttt{n\_optional=2}), with per-transform application probability \texttt{p=0.5}. In addition, fixed transforms are applied every time as listed below. The augmentation strength is ramped during training via a schedule over \texttt{ramp=1000} steps.}
\label{tab:noise_layer}
\begin{tabular}{@{}lll@{}}
\toprule
\textbf{Type} & \textbf{Transform} & \textbf{Low-severity setting / range} \\
\midrule
Fixed & Horizontal flip & Applied with probability $0.5$ \\
Fixed & Resize-to-224 crop & Output size $224\times224$ \\
\midrule
Optional & JPEG compression & JPEG quality sampled in $[70,100]$ (ramped) \\
Optional & Brightness & Factor sampled in $[0.9,1.1]$ (ramped) \\
Optional & Contrast & Factor sampled in $[0.9,1.1]$ (ramped) \\
Optional & Colour jiggle & Brightness $\pm 0.05$, contrast $\pm 0.05$, \\
& & saturation $\pm 0.05$, hue $\pm 0.01$ (ramped) \\
Optional & Grayscale & Applied with probability $0.5$ \\
Optional & Gaussian blur & Kernel size $3$, $\sigma \in [0.1,1.0]$ (ramped) \\
Optional & Gaussian noise & Additive Gaussian noise with std $0.02$ \\
Optional & Hue shift & Hue factor in $[-0.01,0.01]$ (ramped) \\
Optional & Motion blur & Kernel size in $\{3,5\}$, random angle \\
Optional & Posterize & 5 bits \\
Optional & RGB shift & Per-channel shift in $[-0.02,0.02]$ (ramped) \\
Optional & Saturation & Factor sampled in $[0.9,1.1]$ (ramped) \\
Optional & Sharpness & Factor sampled in $[1.0,1.5]$ (ramped) \\
Optional & Median blur & Kernel size $3$ \\
Optional & Box blur & Kernel size $3$ \\
Optional & Perspective distortion & Distortion scale in $[0.05,0.2]$ (ramped) \\
Optional & Resize & Scale sampled in $[0.8,1.2]$, then resized back \\
Optional & Random crop & Crop scale per dimension in $[0.8,1.0]$, then resized back \\
\bottomrule
\end{tabular}
\end{table*}

\begin{table*}[t]
    \caption{Robustness on DIV2K: \BA (\%) under common perturbations.}
  \label{tab:robustness_div2k2}
  \centering
  \scriptsize
  \setlength{\tabcolsep}{4pt}
  \begin{tabular}{@{}lccccccc@{}}
    \toprule
    Attack & TrustMark-Q & InvisMark & PixelSeal & RoSteALS & SSL & MBRS & \textbf{FiLM-CAM} \\
    \midrule
    Crop             & 60.8 & 75.1  & 83.2 & 50.4 & 74.1 & 50.0  & 83.2 \\
    Flip             & 99.9 & 100.0 & 99.6 & 48.7 & 99.9 & 50.0  & 100.0 \\
    Sharpness        & 99.8 & 100.0 & 99.5 & 98.9 & 99.7 & 100.0 & 100.0 \\
    Motion Blur      & 99.8 & 100.0 & 99.6 & 98.7 & 84.5 & 82.2  & 96.9 \\
    Gaussian Noise   & 99.6 & 81.5  & 95.5 & 95.2 & 62.7 & 82.3  & 90.2 \\
    Colour Jiggle    & 99.7 & 100.0 & 99.4 & 98.8 & 99.6 & 99.8  & 99.0 \\
    Brightness       & 99.1 & 99.8  & 99.1 & 98.3 & 99.4 & 99.1  & 99.5 \\
    RGB Shift        & 99.5 & 100.0 & 99.4 & 98.8 & 98.8 & 99.9  & 99.9 \\
    Gamma            & 98.7 & 99.7  & 98.9 & 98.1 & 99.0 & 99.0  & 99.4 \\
    Saturation       & 99.8 & 100.0 & 99.6 & 99.0 & 99.9 & 100.0 & 100.0 \\
    Hue              & 99.8 & 100.0 & 99.5 & 99.0 & 99.9 & 100.0 & 100.0 \\
    Contrast         & 99.0 & 99.9  & 99.1 & 98.6 & 99.3 & 99.6  & 99.8 \\
    Posterize        & 93.6 & 95.9  & 93.4 & 93.4 & 73.0 & 96.0  & 96.0 \\
    Gaussian Blur    & 99.8 & 100.0 & 99.6 & 98.8 & 99.6 & 86.5  & 99.0 \\
    Resize           & 99.8 & 100.0 & 99.6 & 99.0 & 99.8 & 97.9  & 99.9 \\
    Rotate           & 64.6 & 78.9  & 99.0 & 58.1 & 99.1 & 52.0  & 96.0 \\
    Box Blur         & 99.8 & 100.0 & 99.5 & 98.2 & 93.3 & 47.3  & 97.6 \\
    Median Blur      & 99.8 & 100.0 & 99.6 & 98.9 & 99.5 & 97.0  & 99.9 \\
    Grayscale        & 97.0 & 53.4  & 99.6 & 95.8 & 94.9 & 100.0 & 75.7 \\
    JPEG             & 99.8 & 100.0 & 99.5 & 98.6 & 97.4 & 100.0 & 85.1 \\
    Perspective      & 51.2 & 63.8  & 96.0 & 49.7 & 82.4 & 50.1  & 90.2 \\
    \midrule
    Mean             & 93.4 & 92.8  & \textbf{98.0} & 89.2 & 93.1 & 85.2  & \underline{96.0} \\
    \bottomrule
  \end{tabular}
\end{table*}

\section{WAVES Benchmark}
\label{app:waves}
We evaluate the noise-trained FiLM-CAM on the WAVES benchmark \cite{an2024waves}, following its protocol. We report TPR at 0.1\% FPR, bit accuracy, AUC, and the PSNR of the attacked image (\Cref{tab:waves}). FiLM-CAM remains reliably detectable under most single distortions, under photometric composites, and under all adversarial-embedding and surrogate-classifier attacks (TPR $\geq$ 0.98). It is less robust to blur, rotation and JPEG, and to VAE-based regeneration. Diffusion regeneration and rinsing, as well as the geometric and full composite distortions, reduce detection to chance. These attacks destroy the watermark without the key, but do not recover the cover image (attacked PSNR $\leq$ 18\,dB for the diffusion-based attacks).

\begin{table}[t]
  \centering
  \scriptsize
  \setlength{\tabcolsep}{4pt}
  \caption{WAVES benchmark results for FiLM-CAM (noise-trained), with the correct key.}
  \label{tab:waves}
  \begin{tabular}{@{}llcccc@{}}
    \toprule
        Family & Attack & TPR@0.1\%FPR $\uparrow$ & \BA (\%)$\uparrow$ & AUC $\uparrow$ & Attacked PSNR (dB) \\
    \midrule
    Clean & None & 1.000 & 99.66 & 1.000 & 40.56 \\
    \midrule
    \multirow{8}{*}{Distortion (single)}
      & Random crop & 0.995 & 94.1 & 1.000 & 12.86 \\
      & Erase       & 0.997 & 93.0 & 1.000 & 15.66 \\
      & Brightness  & 0.981 & 94.7 & 0.996 & 15.19 \\
      & Contrast    & 0.997 & 95.5 & 1.000 & 21.64 \\
      & Noise       & 0.999 & 95.0 & 1.000 & 26.53 \\
      & JPEG        & 0.742 & 78.7 & 0.952 & 28.89 \\
      & Blur        & 0.429 & 69.5 & 0.763 & 21.22 \\
      & Rotation    & 0.333 & 63.1 & 0.673 & 10.31 \\
    \midrule
    \multirow{4}{*}{Distortion (composite)}
      & Photometric & 0.971 & 90.2 & 0.994 & 13.89 \\
      & Degradation & 0.009 & 51.5 & 0.569 & 20.35 \\
      & Geometric   & 0.000 & 50.3 & 0.517 & 10.54 \\
      & All         & 0.000 & 50.2 & 0.507 & 8.88 \\
    \midrule
    \multirow{4}{*}{Regeneration (VAE)}
      & VAE            & 0.519 & 66.6 & 0.893 & 27.13 \\
      & KL-VAE         & 0.752 & 74.2 & 0.953 & 25.82 \\
      & 4$\times$VAE   & 0.198 & 58.2 & 0.801 & 26.54 \\
      & 4$\times$KL-VAE& 0.619 & 69.1 & 0.926 & 24.25 \\
    \midrule
    \multirow{2}{*}{Regeneration (diffusion)}
      & Diffusion        & 0.000 & 50.2 & 0.507 & 17.73 \\
      & Diffusion-prompt & 0.001 & 50.2 & 0.506 & 16.24 \\
    \midrule
    \multirow{2}{*}{Rinsing (diffusion)}
      & 2$\times$Diffusion & 0.001 & 50.2 & 0.505 & 14.91 \\
      & 4$\times$Diffusion & 0.000 & 50.1 & 0.501 & 13.53 \\
    \midrule
    \multirow{4}{*}{Adversarial}
      & AdvEmb (RN18)       & 1.000 & 99.4 & 1.000 & 35.33 \\
      & AdvCls UnWM\&WM     & 0.991 & 92.5 & 0.999 & 14.40 \\
      & AdvCls Real\&WM     & 0.994 & 93.3 & 0.999 & 14.41 \\
      & AdvCls WM1\&WM2     & 0.995 & 94.6 & 0.999 & 14.40 \\
    \bottomrule
  \end{tabular}
\end{table}

\begin{table*}[t]
  \caption{Clean vs.\ noise-trained FiLM-CAM on COCO, DIV2K and SA-1B.}
  \label{tab:filmcam_clean_vs_noise_appexdix}
  \centering
  \scriptsize
  \setlength{\tabcolsep}{5pt}
  \begin{tabular}{@{}llcccc@{}}
    \toprule
    \multirow{2}{*}{Dataset} & \multirow{2}{*}{Metric} 
    & \multicolumn{2}{c}{FiLM-CAM (Clean)} 
    & \multicolumn{2}{c}{FiLM-CAM (Noise-trained)} \\
    \cmidrule(lr){3-4} \cmidrule(lr){5-6}
        & & $\checkmark$ key & $\times$ key & $\checkmark$ key & $\times$ key \\
    \midrule
    \multirow{4}{*}{COCO}
      & \BA   & 0.99 & 0.53 & 0.99 & 0.55 \\
      & \PSNRW  & 48.2 & --   & 40.2 & --   \\
      &\BAR     & 0.53 & 0.58 & 0.57 & \textbf{0.81} \\
      & \PSNRR    & 59.9 & 56.2 & 57.9 & \textbf{48.0} \\
    \midrule
    \multirow{4}{*}{DIV2K}
      & \BA   & 0.99 & 0.54 & 0.99 & 0.55 \\
      & \PSNRW   & 46.8 & --   & 40.1 & --   \\
      & \BAR     & 0.54 & 0.59& 0.58 & \textbf{0.79} \\
      & \PSNRR      & 55.4 & 53.1 & 54.0 & \textbf{46.6} \\
    \midrule
    \multirow{4}{*}{SA-1B}
      & \BA    & 0.99 & 0.53 & 0.99 & 0.55 \\
      &\PSNRW    & 46.6 & --   & 40.1 & --   \\
      & \BAR     & 0.55& 0.59 & 0.58 & \textbf{0.78 }\\
      & \PSNRR     & 55.3 & 53.1 & 54.0 & \textbf{46.7} \\
    \bottomrule
  \end{tabular}
\end{table*}

\section{Threat Model and Key-Free Attacks}
\label{app:threat_model}

\textbf{Setting.} A single key-conditioned embedder $E$ and extractor $X$ are shared across many users. Each user $u$ owns a distinct secret key $k_u$, and a watermarked image $x_w$ is decoded as $\hat{m} = X(x_w, k_u)$ only with the matching key. We consider an adversary targeting a victim key $k^\star$. It (i) holds images watermarked under $k^\star$, (ii) knows the architecture and weights, (iii) may query the embedding pipeline under $k^\star$ as a black box, with covers and messages of its choice, and (iv) may be a legitimate user holding its own keys $k_a \neq k^\star$. It never observes $k^\star$. Its goals are (a) to decode the hidden message, or (b) to remove the watermark from $k^\star$-marked images while preserving image quality, without learning $k^\star$. We evaluate one attack for each goal below: a black-box surrogate decoder for (a) and a white-box collusion attack for (b).

\textbf{Guarantees and scope.} Without $k^\star$, FiLM-CAM is designed to prevent (a) decoding of the message and (b) \emph{reconstructive} removal that recovers the cover image, and it resists both attacks below. It does not prevent key-free \emph{destruction} of the watermark, e.g.\ by overwriting (Appendix~\ref{app:overwrite}) or strong diffusion regeneration (Appendix~\ref{app:waves}); this is common to learned watermarking methods. We do not evaluate key-recovery attacks that optimize a surrogate key against the white-box extractor; the sharp key-correctness response (\Cref{fig:key_correctness}) suggests partial keys give little advantage, but we leave a full analysis to future work.

\textbf{Surrogate decoder (key-free decoding).} In a black-box, chosen-message setting, the adversary queries the embedding pipeline under $k^\star$ to build a training set of (watermarked image, message) pairs, but never observes $k^\star$ or its FiLM embedding $e_{k^\star}$. It then trains a keyless decoder (ImageNet-pretrained ResNet-50 \cite{he2016deep}, BCE loss) and is evaluated on held-out covers unseen during training. We report the most favourable regime for the attacker, in which a fixed dictionary of 100 messages is reused on new covers (5{,}000 training and 1{,}000 held-out covers, 30 epochs). The surrogate fits the training set (0.998 bit accuracy) but does not generalize: held-out accuracy is 0.52, at chance level (\Cref{tab:surrogate}).

\begin{table}[t]
  \centering
  \scriptsize
  \setlength{\tabcolsep}{6pt}
  \caption{Key-free decoding with a surrogate decoder, on held-out covers.}
  \label{tab:surrogate}
  \begin{tabular}{@{}lc@{}}
    \toprule
    Decoder & \BA (held-out) \\
    \midrule
    FiLM-CAM extractor, correct key & 0.99 \\
    FiLM-CAM extractor, random key  & 0.54 \\
    Surrogate decoder (no key)      & 0.52 \\
    \bottomrule
  \end{tabular}
\end{table}

\textbf{Collusion (key-free removal).} A white-box adversary collects a set of images watermarked under $k^\star$, each with a different cover and a fresh random message. Since it chose the covers, it can compute their pixel-space watermark residuals. It averages these into a single perturbation and subtracts it, at scale $\lambda$, from a disjoint set of images watermarked under $k^\star$. Successful removal would drive bit accuracy toward 50\% while preserving image quality. As \Cref{tab:collusion} shows, bit accuracy remains $\geq$90\% across the quality-preserving range ($\lambda \leq 2$, PSNR $\approx$ 40\,dB, matching the un-attacked watermarked image). It falls toward chance only at large $\lambda$, where PSNR drops below 30\,dB with visible artifacts. Pooling same-key images therefore does not yield a perturbation that strips the watermark while preserving the image.

\begin{table}[t]
  \centering
  \scriptsize
  \setlength{\tabcolsep}{6pt}
  \caption{Collusion attack: subtracting the averaged same-key watermark at scale $\lambda$.}
  \label{tab:collusion}
  \begin{tabular}{@{}ccc@{}}
    \toprule
        $\lambda$ & \BA (\%) & PSNR (dB) \\
    \midrule
    0.5  & 98.87 & 40.57 \\
    1.0  & 97.59 & 40.71 \\
    2.0  & 90.48 & 40.14 \\
    3.0  & 80.71 & 38.76 \\
    5.0  & 67.97 & 35.49 \\
    8.0  & 58.55 & 31.58 \\
    10.0 & 55.93 & 29.63 \\
    15.0 & 52.19 & 26.08 \\
    \bottomrule
  \end{tabular}
\end{table}

\section{Embedder/Extractor Architecture Details}
\paragraph{\textbf{Embedder.}}
The watermark embedder takes as input a cover image
$x \in \mathbb{R}^{3 \times 256 \times 256}$,
a binary secret
$m \in \{0,1\}^{100}$,
and a binary access key
$k \in \{0,1\}^{100}$,
and outputs a watermark residual
$r \in \mathbb{R}^{3 \times 256 \times 256}$,
which is composed with the cover image to obtain the watermarked image
$x^{\mathrm{w}}$.
The embedder is based on a lightweight U-Net.
Unlike bottleneck-based message injection schemes, we first map the secret ($m$) to a spatial representation through a \texttt{Secret2Image} module, which projects the 100-bit secret to a
$3 \times 16 \times 16$ tensor and upsamples it to
$3 \times 256 \times 256$.
This secret image is concatenated with the RGB cover image, producing a 6-channel input.
The encoder begins with a residual block and three downsampling blocks, producing feature maps of sizes
$32 \times 256 \times 256$,
$64 \times 128 \times 128$,
$128 \times 64 \times 64$,
and
$256 \times 32 \times 32$.
The bottleneck consists of eight residual blocks operating at
$32 \times 32$
resolution.
To enforce conditional access, the binary key is mapped by a three-layer MLP to a 64-dimensional key embedding, which is transformed into FiLM parameters
$(\gamma,\beta) \in \mathbb{R}^{256}$ and applied channel-wise to the bottleneck features.
The decoder mirrors the encoder with three bilinear upsampling blocks and skip connections, and a final
$1 \times 1$
convolution predicts the watermark residual.

\paragraph{\textbf{Extractor.}}
The extractor takes as input a watermarked image
$x^{\mathrm{w}}$
and the access key
$k$,
and predicts secret logits
$\tilde{m} \in \mathbb{R}^{100}$,
which are thresholded at inference time to recover the binary message
$\hat{m} \in \{0,1\}^{100}$.
In our clean implementation, the extractor is based on a ConvNeXt-Base \cite{liu2022convnet} backbone.
The input image is first resized to
$224 \times 224$,
then processed by ConvNeXt-Base to obtain a high-level feature tensor.
Global average pooling and BatchNorm reduce this representation to a 1024-dimensional descriptor.

The same key-conditioning branch used in the embedder maps the 100-bit key to a 64-dimensional embedding, from which a FiLM generator predicts
$(\gamma,\beta) \in \mathbb{R}^{1024}$.
These parameters modulate the pooled image descriptor before a final linear layer predicts the 100 secret logits.
This design ensures that correct decoding depends on presenting the correct key, while incorrect keys perturb the feature representation and degrade extraction.
\begin{table}[t]
  \caption{High-level architecture of the FiLM-CAM watermark embedder.}
  \label{tab:embedder_arch}
  \centering
  \scriptsize
  \setlength{\tabcolsep}{5pt}
  \begin{tabular}{@{}l c@{}}
    \toprule
    \multicolumn{2}{c}{\textbf{Embedder}} \\
    \midrule
    Input &
    $x \in \mathbb{R}^{3\times256\times256},\;
    m \in \{0,1\}^{100},\;
    k \in \{0,1\}^{100}$ \\
    Secret projection &
    Linear, reshape, upsample:
    $m \rightarrow \mathbb{R}^{3\times16\times16} \rightarrow \mathbb{R}^{3\times256\times256}$ \\
    Concatenation &
    $\mathrm{Concat}(x,\mathrm{Secret2Image}(m))
    \rightarrow \mathbb{R}^{6\times256\times256}$ \\
    Initial block &
    ResNetBlock $\rightarrow \mathbb{R}^{32\times256\times256}$ \\
    Down block 1 &
    DBlock $\rightarrow \mathbb{R}^{64\times128\times128}$ \\
    Down block 2 &
    DBlock $\rightarrow \mathbb{R}^{128\times64\times64}$ \\
    Down block 3 &
    DBlock $\rightarrow \mathbb{R}^{256\times32\times32}$ \\
    Bottleneck &
    $8 \times$ ResNetBlock $\rightarrow \mathbb{R}^{256\times32\times32}$ \\
    Key embedding &
    MLP:
    $k \rightarrow e_k \in \mathbb{R}^{64}$ \\
    FiLM generation &
    Linear:
    $e_k \rightarrow (\gamma,\beta),\;
    \gamma,\beta \in \mathbb{R}^{256}$ \\
    Key conditioning &
    $\mathrm{FiLM}(z;\gamma,\beta)
    \rightarrow \mathbb{R}^{256\times32\times32}$ \\
    Up block 1 &
    Skip concat, UBlock $\rightarrow \mathbb{R}^{128\times64\times64}$ \\
    Up block 2 &
    Skip concat, UBlock $\rightarrow \mathbb{R}^{64\times128\times128}$ \\
    Up block 3 &
    Skip concat, UBlock $\rightarrow \mathbb{R}^{32\times256\times256}$ \\
    Output layer &
    Conv\(1\times1\) $\rightarrow r \in \mathbb{R}^{3\times256\times256}$ \\
    Watermark composition &
    $r$ composed with $x \rightarrow x^{\mathrm{w}} \in \mathbb{R}^{3\times256\times256}$ \\
    \bottomrule
  \end{tabular}
\end{table}

\begin{table}[t]
  \caption{High-level architecture of the FiLM-CAM watermark extractor.}
  \label{tab:extractor_arch}
  \centering
  \scriptsize
  \setlength{\tabcolsep}{5pt}
  \begin{tabular}{@{}l c@{}}
    \toprule
    \multicolumn{2}{c}{\textbf{Extractor}} \\
    \midrule
    Input &
    $x^{\mathrm{w}} \in \mathbb{R}^{3\times H\times W},\;
    k \in \{0,1\}^{100}$ \\
    Resize &
    Bilinear interpolation $\rightarrow \mathbb{R}^{3\times224\times224}$ \\
    Image encoder &
    ConvNeXt-Base features $\rightarrow \mathbb{R}^{1024\times7\times7}$ \\
    Global pooling &
    Adaptive average pooling $\rightarrow \mathbb{R}^{1024\times1\times1}$ \\
    Normalization &
    BatchNorm2D $\rightarrow \mathbb{R}^{1024\times1\times1}$ \\
    Flatten &
    $\rightarrow \mathbb{R}^{1024}$ \\
    Key embedding &
    MLP:
    $k \rightarrow e_k \in \mathbb{R}^{64}$ \\
    FiLM generation &
    Linear:
    $e_k \rightarrow (\gamma,\beta),\;
    \gamma,\beta \in \mathbb{R}^{1024}$ \\
    Key conditioning &
    $\mathrm{FiLM}(z;\gamma,\beta)
    \rightarrow \mathbb{R}^{1024}$ \\
    Bit predictor &
    Linear $\rightarrow \tilde{m} \in \mathbb{R}^{100}$ \\
    Inference &
    Sigmoid / threshold $\rightarrow \hat{m} \in \{0,1\}^{100}$ \\
    \bottomrule
  \end{tabular}
\end{table}

\begin{figure}[t!]
  \centering
  \includegraphics[width=0.95\linewidth,height=15cm]{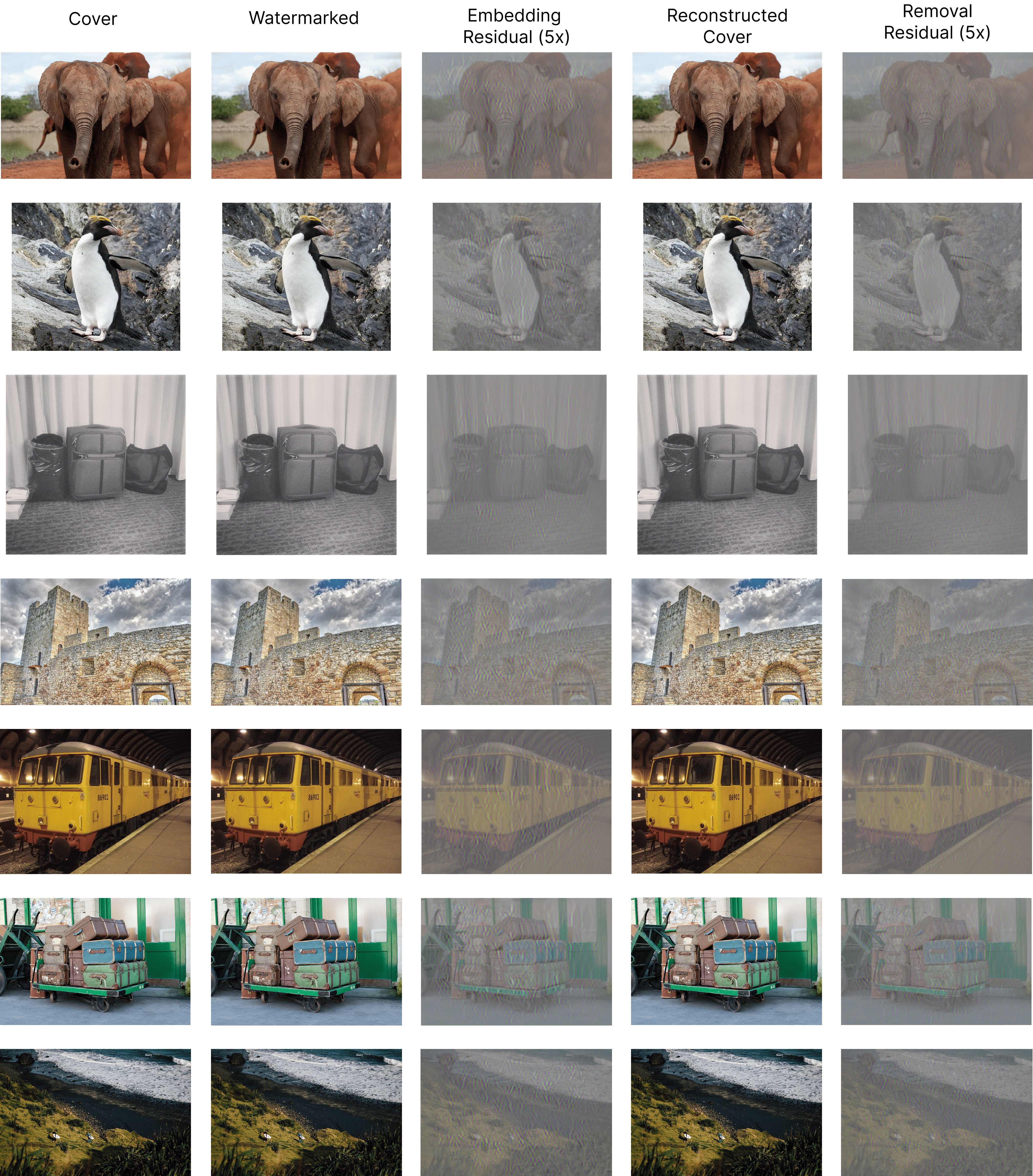}
  \caption{Qualitative results of embedding and removal with FiLM-CAM. From left to right, we show the cover image, the watermarked image, the embedding residual ($5\times$ amplified for visibility), the reconstructed cover after removal, and the removal residual ($5\times$ amplified).}
  \label{fig:film_cam_wm_results}
\end{figure}

\section{FiLM Embedding Dimension}

We ablate the dimensionality of the FiLM key embedding while keeping the rest of FiLM-CAM fixed. \cref{tab:film_embedding_dim} compares the default 64-dimensional setting against larger embeddings of 100 and 200 dimensions. Overall, the results show that the method is not highly sensitive to the exact embedding size, and all three settings achieve similar image quality and conditional behaviour.

The default 64-dimensional embedding gives the best correct-key extraction accuracy, reaching $0.99$, while maintaining a wrong-key accuracy of $0.54$, close to random guessing. Increasing the embedding size to 100 dimensions yields very similar performance, with a small drop in correct-key accuracy to $0.98$ and a wrong-key accuracy of $0.55$. Using a 200-dimensional embedding slightly improves PSNR and gives the lowest wrong-key accuracy at $0.52$, but also reduces correct-key extraction accuracy to $0.96$.

Overall, these results suggest a mild trade-off between watermark fidelity, correct-key decoding, and wrong-key collapse as the FiLM embedding dimension increases. We therefore use 64 dimensions as the default setting, since it provides the best overall balance and achieves the highest correct-key extraction accuracy.

\begin{table}[t]
  \centering
  \scriptsize
  \setlength{\tabcolsep}{6pt}
    \caption{Ablation of the FiLM key-embedding dimension (default 64).}
  \label{tab:film_embedding_dim}
  \begin{tabular}{@{}lccc@{}}
    \toprule
        \textbf{FiLM emb. dim.} & \textbf{\PSNRW$\uparrow$} & \textbf{\BA ($\checkmark$)$\uparrow$} & \textbf{\BA ($\times$)} \\
    \midrule
    100 & 46.6 & 0.98 & 0.55 \\
    200 & \textbf{46.8} & 0.96 & \textbf{0.52} \\
    \midrule
    \textbf{64 (default)} & 46.7 & \textbf{0.99} & 0.54 \\
    \bottomrule
  \end{tabular}
\end{table}

\section{Additional Key-Conditional Analysis}

\begin{figure}[t!]
  \centering
  \includegraphics[width=0.95\linewidth,height=17cm]{images/residual_comparison_2.jpg}
  \caption{Effect of Different Keys on the Watermark Residuals}
  \label{fig:residual_comparison}
\end{figure}
\textbf{Effect of the Keys on the Watermark Residual.} To further study the role of the access key in the embedder, we perform a controlled experiment in which the cover image and secret are kept fixed, while only the key is changed. Specifically, for the same input image $x$ and secret $m$, we generate two watermarked images using two different keys, $k_A$ and $k_B$, producing corresponding residuals $r_A$ and $r_B$.

\cref{fig:residual_comparison} shows that the residuals obtained with different keys are visually different when magnified, even though the cover image and secret are identical. We also visualize the residual-difference image $r_A-r_B$, which further highlights the structured differences induced solely by changing the key. This provides qualitative evidence that the access key influences the actual embedding pattern, rather than acting only at decoding time.

We additionally perform a decoding cross-check. A watermarked image generated with key $k_A$ decodes correctly with $k_A$ at $98.0\%$ bit accuracy, but drops to $49.0\%$ when decoded with the wrong key $k_B$. Likewise, a watermarked image generated with key $k_B$ achieves $100.0\%$ bit accuracy with $k_B$ and $49.0\%$ with the wrong key $k_A$. Since random guessing corresponds to $50\%$, these results confirm that the embeddings produced by different keys are key-specific and are not interchangeable at decoding time.

Quantitatively, the mean absolute error (MAE) between the amplified residual visualizations for the two keys is $0.0875$, while the corresponding amplified MAE between the watermarked images is $0.0727$. Although the final watermarked images remain visually similar, the underlying residuals differ in a consistent and key-dependent manner.

\noindent\textbf{Effect of Wrong Keys on Removal.} We also examine removal behaviour across different incorrect keys. Starting from a watermarked image generated with secret $m$ and key $k_A$, we attempt removal using the correct key $k_A$ and several wrong keys $k_B,k_C,k_D$. Removal with the correct key achieves a PSNR of approximately $55$ dB, whereas removal with the wrong keys consistently gives much lower PSNR values of around $49$ dB. In \cref{fig:residual_comparison}, we visualize the corresponding removal residuals and the residual-difference maps between the correct-key and wrong-key cases. The residual patterns for different wrong keys appear broadly similar, which is consistent with their similarly low PSNR values. This suggests that once the correct key is absent, the remover produces residuals in a consistent manner across different incorrect keys.

\section{Removal under Fully Unauthorized Queries}
\label{app:removal_inputs}
In the main paper, removal is analyzed primarily as a function of key correctness, showing that FiLM-CAM achieves high-quality removal under the correct key while removal with incorrect keys remains much less effective. Since FiLM-CAM removal is conditioned on the image, secret, and key, we additionally study a stronger unauthorized setting in which the attacker does not possess either the correct key or the correct secret. This setting is practically relevant, since an unauthorized user would typically lack both pieces of information.

In \cref{tab:unauthorized_removal} we compare two cases across COCO, DIV2K, and SA-1B: (i) removal with an incorrect key while keeping the secret fixed (correct), and (ii) removal with both a random secret and a random key. We report post-removal watermark bit accuracy, where values near $0.50$ indicate successful removal and higher values indicate that more watermark information remains after the attempted removal. In the first case, the post-removal bit accuracy stays close to random guessing, around $0.59$, consistent with the wrong-key behavior observed in the main paper. In contrast, when both the secret and key are random, the post-removal bit accuracy rises to approximately $0.81$ across all three datasets. This indicates that in the fully unauthorized setting, FiLM-CAM retains a substantial portion of the watermark signal and becomes significantly harder to remove. 

These results further support the conditional-access property of FiLM-CAM. Unauthorized removal is already limited when the key is incorrect, and becomes substantially less effective when the attacker also lacks the correct secret. In other words, the removal mechanism is not only key-dependent, but also sensitive to the secret supplied to the residual estimator, making successful removal difficult without access to the full authorized query (correct secret + correct key).

\noindent\textbf{Role of the message at removal.} \Cref{tab:removal_message} isolates the effect of the message supplied at removal time, for both correct and random keys. With the correct key, supplying the correct message yields higher removal PSNR (55.66 vs.\ 48.31\,dB) and more complete removal (post-removal bit accuracy 0.57 vs.\ 0.78). With a random key, the message has little effect: removal fails either way, and post-removal bit accuracy stays at 0.81--0.83, i.e.\ the watermark persists. High-fidelity removal therefore requires both the correct key and the correct message.

\begin{table}[t]
  \centering
  \scriptsize
  \setlength{\tabcolsep}{6pt}
    \caption{Effect of the key and message supplied at removal.}
  \label{tab:removal_message}
  \begin{tabular}{@{}llccc@{}}
    \toprule
       Key & Message & \PSNRR$\uparrow$ & \dPSNR$\uparrow$ & \BAR \\
    \midrule
    Correct & Correct & \textbf{55.66} & \textbf{15.58} & \textbf{0.57} \\
    Correct & Random  & 48.31 & 8.23 & 0.78 \\
    \midrule
    Random  & Correct & 46.37 & 6.29 & 0.83 \\
    Random  & Random  & 45.27 & 5.19 & 0.81 \\
    \bottomrule
  \end{tabular}
\end{table}

\begin{table}[t]
  \centering
  \scriptsize
  \setlength{\tabcolsep}{6pt}
    \caption{\BAR of FiLM-CAM (clean) under unauthorized removal; higher means the watermark survives.}
  \label{tab:unauthorized_removal}
  \begin{tabular}{@{}lcc@{}}
    \toprule
    \textbf{Dataset} & \textbf{Fixed Secret + Random Key} & \textbf{Random Secret + Random Key} \\
    \midrule
    COCO  & 0.58 & \textbf{0.81} \\
    DIV2K & 0.59 & \textbf{0.82} \\
    SA-1B & 0.59 & \textbf{0.81} \\
    \bottomrule
  \end{tabular}
\end{table}

\section{Practical Considerations for Removal}

Many regeneration-based watermark removal baselines operate at a fixed internal resolution and output images at that resolution, which can be suboptimal when evaluating fidelity at the original image scale. FiLM-CAM removal can be applied directly at the native resolution via the same scaling protocol used for inference, making it practical for high-resolution inputs. We additionally report removal inference time in the main paper. FiLM-CAM requires only a forward pass through the embedder plus a small number of refinement steps, making it computationally efficient relative to iterative generative baselines.

\begin{table*}[t!]
    \caption{Quality of watermark-removed images w.r.t.\ the original cover.}
  \label{tab:removal_quality_supmat}
  \centering
  \setlength{\tabcolsep}{3pt}
  \footnotesize
  \begin{tabular}{@{}lccc|ccc|ccc@{}}
    \toprule
    & \multicolumn{3}{c|}{COCO} & \multicolumn{3}{c|}{DIV2K} & \multicolumn{3}{c}{SA-1B} \\
    Method
            & \PSNRR$\uparrow$ & SSIM$\uparrow$ & LPIPS$\downarrow$
      & \PSNRR$\uparrow$ & SSIM$\uparrow$ & LPIPS$\downarrow$
      & \PSNRR$\uparrow$ & SSIM$\uparrow$ & LPIPS$\downarrow$ \\
    \midrule
    TrustMark-RM (Q)
      & 41.7 & 0.9900 & 0.0023
      & 40.8 & 0.9936 & 0.0034
      & 41.1 & 0.9940 & 0.0037 \\
    CtrlRegen+
      & 21.1 & 0.6418 & 0.1819
      & 20.0 & 0.5744 & 0.1992
      & 20.4 & 0.6122 & 0.1884 \\
    Diffusion Attack
      & 21.0 & 0.5619 & 0.1816
      & 22.4 & 0.6250 & 0.1513
      & 22.0 & 0.6192 & 0.1575 \\
    VAE-Bmshj2018
      & 30.0 & 0.8399 & 0.2096
      & 30.85 & 0.8961 & 0.1089
      & 29.56 & 0.8661 & 0.1679 \\
    VAE-Cheng2020
      & 31.4 & 0.8661 & 0.1625
      & 30.85 & 0.8961 & 0.1089
      & 31.18 & 0.8990 & 0.1079 \\
    \midrule
    \textbf{FiLM-CAM} ($\checkmark$ key)
      & \textbf{59.9} & \textbf{0.9997} & \textbf{0.0002}
      & \textbf{55.6} & \textbf{0.9991} & \textbf{0.0004}
      & \textbf{55.3} & \textbf{0.9990} & \textbf{0.0004} \\
    \textbf{FiLM-CAM} ($\times$ key)
      & 56.2 & 0.9990 & 0.0004
      & 53.1 & 0.9986 & 0.0014
      & 53.1 & 0.9986 & 0.0015 \\

    \midrule
    \textbf{FiLM-CAM (noise)} ($\checkmark$ key)
      & \underline{57.9} & \underline{0.9996} & \underline{0.0003}
      & \underline{54.0} & \underline{0.9990} & \underline{0.0011}
      & \underline{54.0} & \underline{0.9989}& \underline{0.0011}\\
    \textbf{FiLM-CAM (noise) } ($\times$ key)
      & 48.0 & 0.9960 & 0.0040
      & 46.6 & 0.9953 & 0.0074
      & 46.7 & 0.9953& 0.0074 \\
    \bottomrule
  \end{tabular}
\end{table*}

\section{Visual Comparison with Baselines}
\label{app:visual_comparison}
\begin{figure}[t!]
  \centering
  \includegraphics[width=0.95\linewidth,keepaspectratio]{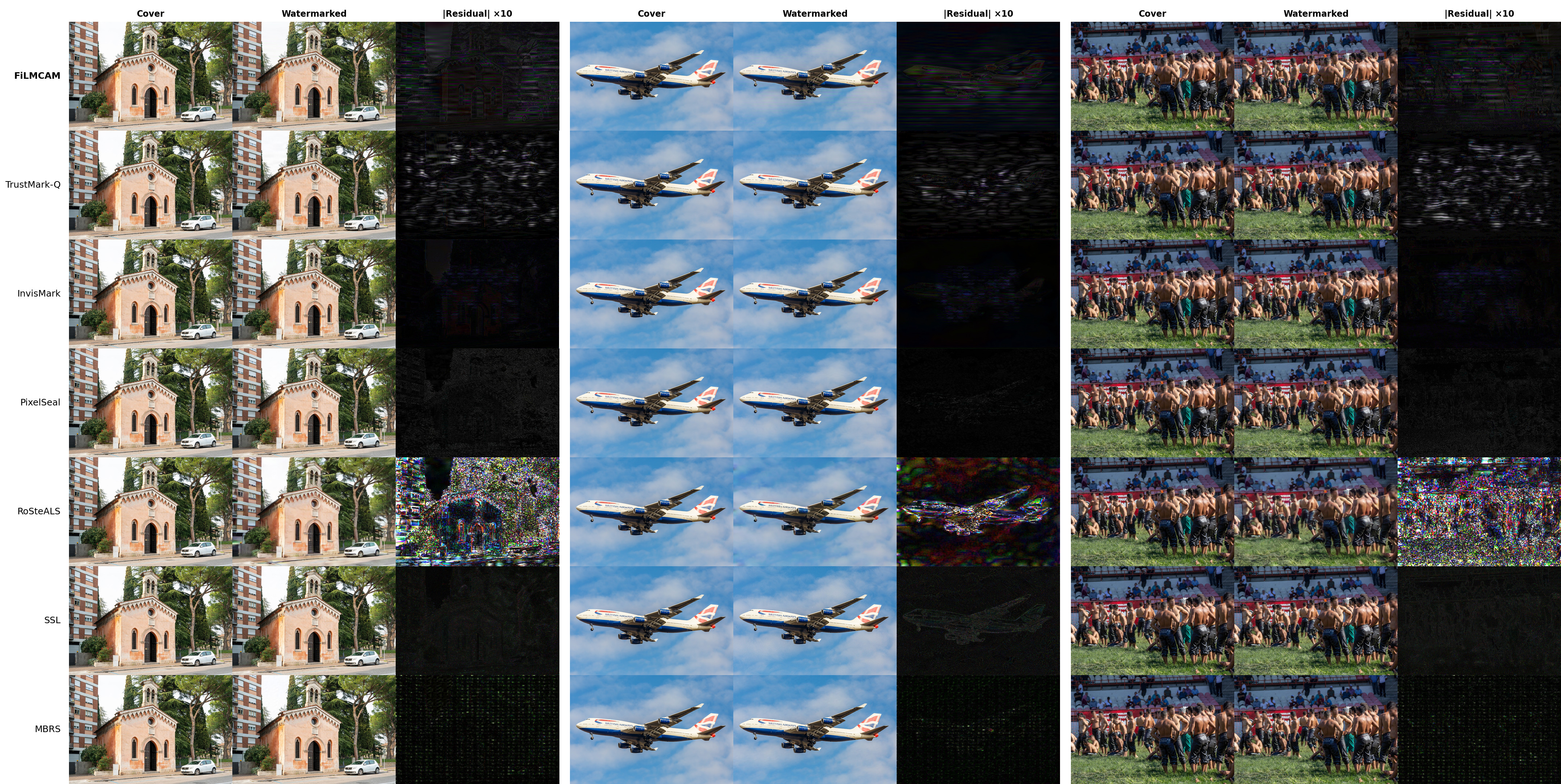}
  \caption{Side-by-side comparison of watermarked images from FiLM-CAM and baseline methods.}
  \label{fig:comparison}
\end{figure}

\Cref{fig:comparison} shows watermarked images produced by FiLM-CAM and the baseline methods for the same cover images, together with the amplified residuals.

\section{Key-Gated Removal without the Noise Layer}
\label{app:finetune}
To isolate the role of the noise layer (\Cref{sec:sched}), we fine-tune the clean FiLM-CAM model using the remover in place of the noise layer: the wrong-key removal residual is subtracted from the watermarked image before decoding, and the extractor is trained with a BCE objective to still recover the true secret under the correct key. \Cref{tab:finetune_remover} reports results on COCO (1K images). After fine-tuning, wrong-key removal leaves the watermark decodable (post-removal bit accuracy 0.94, compared with 0.58 for the clean model in \Cref{tab:filmcam_clean_vs_noise_appexdix}), while correct-key removal still erases it (0.60) at high fidelity (63.22\,dB). This confirms that FiLM provides key-dependent residuals, and a robustly trained extractor turns this into a gap between authorized and unauthorized removal.

\begin{table}[t]
  \centering
  \scriptsize
  \setlength{\tabcolsep}{6pt}
  \caption{FiLM-CAM (clean) fine-tuned with the remover in place of the noise layer, on COCO (1K images).}
  \label{tab:finetune_remover}
  \begin{tabular}{@{}lcccc@{}}
    \toprule
        Setting & \PSNRW & \BA & \PSNRR & \BAR \\
    \midrule
    Correct key & 44.96 & 0.99 & 63.22 & 0.60 \\
    Wrong key   & --    & 0.52 & 55.61 & 0.94 \\
    \bottomrule
  \end{tabular}
\end{table}

\section{Overwriting by Unauthorized Re-Embedding}
\label{app:overwrite}
We test whether an attacker without the correct key can disable an existing watermark by re-embedding over it. On 1K COCO images, we embed watermark $w_1$ with key $k_1$ (giving $\mathrm{wm}_1$), then re-embed a second watermark $w_2$ with a different key $k_2$ (giving $\mathrm{wm}_2$), and decode $w_1$ from both. For the baselines, the second embedding uses a different message. As shown in \Cref{tab:overwrite}, the victim watermark $w_1$ drops to near chance after a single overwrite for FiLM-CAM and for all baselines, while PSNR remains above 39\,dB. Overwriting is therefore a key-free \emph{destruction} attack shared by learned watermarking methods: it disables detection of $w_1$, but does not recover the cover image, leaving a doubly watermarked image with additional distortion. FiLM-CAM's key gating applies to reconstructive removal and does not prevent such destruction.

\begin{table}[t]
  \centering
  \scriptsize
  \setlength{\tabcolsep}{6pt}
  \caption{Overwrite attack on COCO (1K images). $w_1$@$\mathrm{wm}_i$: bit accuracy of the original watermark $w_1$ decoded from $\mathrm{wm}_i$.}
  \label{tab:overwrite}
  \begin{tabular}{@{}lcccc@{}}
    \toprule
    Method & PSNR $\mathrm{wm}_1$ (dB) & PSNR $\mathrm{wm}_2$ (dB) & $w_1$@$\mathrm{wm}_1$ & $w_1$@$\mathrm{wm}_2$ \\
    \midrule
    PixelSeal \cite{souvcek2025pixel}   & 45.90 & 42.94 & 0.99 & 0.50 \\
    TrustMark \cite{bui2025trustmark}   & 41.94 & 39.12 & 0.99 & 0.50 \\
    InvisMark \cite{xu2024invismark}    & 48.36 & 44.72 & 0.99 & 0.54 \\
    FiLM-CAM (ours)                     & 48.57 & 43.29 & 0.99 & 0.53 \\
    \bottomrule
  \end{tabular}
\end{table}

\clearpage

\newpage

\end{document}